\documentclass[10pt,twocolumn,letterpaper]{article}

\usepackage[pagenumbers]{cvpr} 

\usepackage{bbm}
\usepackage{multirow}
\usepackage{color, colortbl}
\usepackage{pifont}
\usepackage{makecell}
\usepackage{ragged2e}
\usepackage{nicematrix}
\usepackage{booktabs}
\usepackage[accsupp]{axessibility}  

\definecolor{myblue2}{RGB}{198,213,250}
\definecolor{Gray}{gray}{0.95}
\definecolor{Gray8}{gray}{0.85}
\definecolor{Gray7}{gray}{0.75}

\newcommand{\cmark}{\ding{51}}%
\newcommand{\xmark}{\ding{55}}%
\definecolor{cvprblue}{rgb}{0.21,0.49,0.74}
\usepackage[pagebackref,breaklinks,colorlinks,allcolors=cvprblue]{hyperref}

\def\paperID{273} 
\def\confName{CVPR}
\def\confYear{2025}

\title{UniPre3D: Unified Pre-training of 3D Point Cloud Models with \\ Cross-Modal Gaussian Splatting}

\author{
    Ziyi Wang\thanks{Equal contribution. ~\textsuperscript{\dag}Corresponding author.} ~~~~
    Yanran Zhang$^*$ ~~~
    Jie Zhou ~~~
    Jiwen Lu$^{\dagger}$      \\
    Department of Automation, Tsinghua University, China\\
    {\tt\small \{wziyi22, zhangyr21\}@mails.tsinghua.edu.cn;} {\tt\small \{jzhou, lujiwen\}@tsinghua.edu.cn} \\
}

\begin{document}
\maketitle
\begin{abstract}
The scale diversity of point cloud data presents significant challenges in developing unified representation learning techniques for 3D vision. Currently, there are few unified 3D models, and no existing pre-training method is equally effective for both object- and scene-level point clouds. In this paper, we introduce UniPre3D, the first unified pre-training method that can be seamlessly applied to point clouds of any scale and 3D models of any architecture. Our approach predicts Gaussian primitives as the pre-training task and employs differentiable Gaussian splatting to render images, enabling precise pixel-level supervision and end-to-end optimization. To further regulate the complexity of the pre-training task and direct the model's focus toward geometric structures, we integrate 2D features from pre-trained image models to incorporate well-established texture knowledge. We validate the universal effectiveness of our proposed method through extensive experiments across a variety of object- and scene-level tasks, using diverse point cloud models as backbones. Code is available at \url{https://github.com/wangzy22/UniPre3D}.
\end{abstract}    
\section{Introduction}

Recently, the unification of model architectures and learning mechanisms has become a prominent research focus, as it represents a crucial milestone toward achieving Artificial General Intelligence. Significant progress has been made in the 2D vision domain, where unified models~\cite{lu2022unifiedio, zhu2022uniperceiver, zhang2023metatransformer, wang2023beitv3} have been developed for multi-modal data. However, in the 3D domain, only a few unified models~\cite{zhou2023uni3d, chen2023unit3d} have been introduced, and their impact has not been as substantial as that of unified image models. A key challenge lies in the greater scale diversity of point clouds compared to images. Images generally have a similar number of pixels and information density, whether depicting a single object or a complex scene. In contrast, scene-level point clouds can contain hundreds of times more points than object-level point clouds. As a result, point-based models~\cite{qi2017pointnet, qi2017pointnet2}, which excel at capturing fine-grained local structures in object data, often struggle to handle large-scale scene samples effectively. Conversely, voxel-based models~\cite{choy2019spunet, peng2024oacnn}, while adept at capturing long-range global relations in scene data, tend to lose geometric details when processing object samples. 

\begin{figure}[t]
  \centering
  \includegraphics[width=\linewidth]{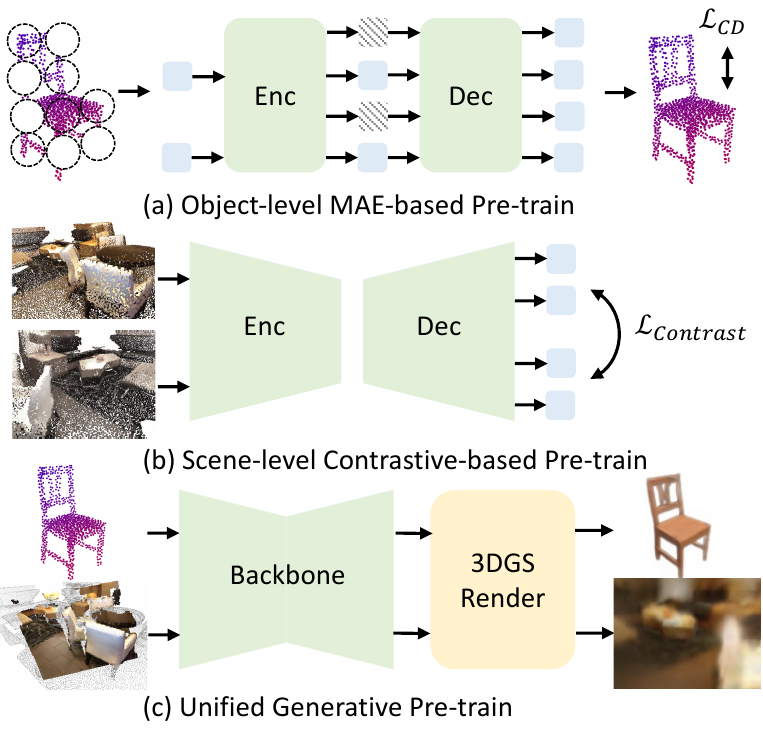}
  \vspace{-20pt}
  \caption{\textbf{Pre-training paradigm comparison.} Existing object-level pre-training methods usually follow a generative masked auto-encoding (MAE) paradigm. Their scene-level counterparts mostly leverage the contrastive learning paradigm. We propose a unified pre-training method that is applicable and effective to both object- and scene-level point clouds and models.}
   \label{fig:overview}
   \vspace{-10pt}
\end{figure}

Given the distinct representation learning paradigms for object and scene point clouds, existing 3D pre-training methods are naturally divided into two streams based on data scale, illustrated in Figure~\ref{fig:overview}. For object point clouds, generative masked auto-encoding (MAE)~\cite{pang2022pointmae, he2022mae} has been widely adopted. However, this approach proves ineffective for scene samples, where the set-to-set Chamfer distance loss is computationally expensive and fails to supervise large-scale data. For scene point clouds, contrastive learning~\cite{xie2020pc, chen2020simclr} forms the foundation of most pre-training methods, as the complexity of scene samples and data augmentations make contrastive tasks challenging enough to serve as effective pre-training. Unfortunately, for comparatively simple object data, contrastive learning tends to saturate quickly, limiting its effectiveness. Currently, there is no unified pre-training method in the 3D domain that is robust to the scale diversity of point clouds.

In this paper, we propose UniPre3D, the first unified pre-training method for the 3D domain that accommodates point clouds of varying scales and 3D models with diverse structures. Our approach centers on predicting Gaussian primitives, which can be rendered into view images via differentiable Gaussian splatting~\cite{kerbl20233dgs}. This enables end-to-end optimization and allows for precise pixel-wise supervision in the image domain. During the pre-training, the 3D model is encouraged to learn local structures that capture fine-grained details like color and opacity to produce more realistic rendered images. The model is also optimized to build global relations that adjust Gaussian positions and covariances to achieve overall coordination. To further control the complexity of the pre-training task, we propose scale-adaptive fusion techniques. We integrate 2D features from pre-trained image models with 3D features from the backbone model, supplementing extra color and texture knowledge to enhance the model's focus on geometry. Since the scale complexity of the projected images aligns with that of the input point cloud, and due to the inherent flexibility of Gaussian primitives, UniPre3D is self-adaptive to both object- and scene-level point clouds and the pre-training task complexity is effectively balanced.

We conduct extensive experiments to validate the unified effectiveness of UniPre3D, including classification on the ScanObjectNN~\cite{Uy2019scanobjectnn} and part segmentation on the ShapeNetPart~\cite{yi2016shapenetpart}, as well as semantic and instance segmentation on scene-level datasets such as ScanNet20~\cite{dai2017scannet}, ScanNet200~\cite{rozenberszki2022scannet200}, and S3DIS~\cite{armeni2016s3dis}. For both object- and scene-level experiments, we select at least one standard model and one advanced model as the backbone to demonstrate that UniPre3D consistently improves performance across various point cloud models, highlighting its architecture-agnostic nature. UniPre3D consistently outperforms previous methods under most benchmarks.

In conclusion, the contributions of our paper are as follows: (1) We propose UniPre3D, the first unified pre-training method for point clouds of any scale and 3D models of any architecture. (2) We propose scale-adaptive fusion techniques to integrate pre-trained image features with 3D features, effectively controlling the pre-training task complexity. Both are verified via our extensive experiments across various 3D perception tasks and diverse backbones.

\section{Related Work}

\noindent\textbf{Point Cloud Perception.} To tackle the unordered and sparse structure of point clouds, two primary approaches for representation learning have emerged: point-based and voxel-based. Point-based methods use K-nearest neighbors (KNN) or ball query algorithms to define neighborhood regions, followed by various mechanisms to aggregate local features and model global relationships. This approach is more commonly used in object-level perception models~\cite{qi2017pointnet, qi2017pointnet2, qian2022pointnext, pointmlp, chen2023dela, park2023spotr, wang2024gpsformer, han2024mamba3d, zhang2024pcm, feng2024interpretable3d} that prioritize capturing fine-grained local structures. In contrast, voxel-based methods first convert the sparse point cloud into sparse voxels, and then apply efficient sparse convolution for feature extraction. This approach is often employed in scene-level perception models~\cite{choy2019spunet, peng2024oacnn, kolodiazhnyi2024oneformer3d}, since scene point clouds are more complex and long-range correlations are crucial to large-scale perception. Recently, Transformer-based models~\cite{lai2022st, wu2022ptv2, zhao2021pt, yang2023swin3d, thomas2024kpconvx} and serialized architectures~\cite{wang2023octformer, wu2024ptv3} have also been introduced to improve the construction of global structures in 3D scene perception.

\vspace{6pt}
\noindent\textbf{Object-level Pre-training.} Since the introduction of masked auto-encoding~\cite{he2022mae} into 3D object pre-training by Point-MAE~\cite{pang2022pointmae}, subsequent models~\cite{liu2022maskpoint, zhang2023i2pmae, zhang2024pcpmae, ren2024pointcmae, zha2024pointfemae, liu2023pointrae, zhang2022pointm2ae} have developed various strategies to enhance pre-training effectiveness. They primarily mask portions of the point cloud and use the Transformer attention to infer the masked regions. Another line of research~\cite{dong2022act, wang2023jm3d, qi2023recon, qi2025recon++, xue2023ulip, xue2024ulip2} incorporates large language models, pre-trained image models, or both, to enable multimodal pre-training. They leverage the extensive pre-trained knowledge to achieve superior fine-tuning performance. Recently, generative pre-training~\cite{wang2023tap, chen2024pointgpt, qi2024vpp, zheng2024pointdif} has emerged as a competitive paradigm for object pre-training, focusing on conditional generation as the pre-training task. Our proposed UniPre3D pipeline adopts this generative pre-training paradigm, while delivering improved efficiency and broader applicability.

\vspace{6pt}
\noindent\textbf{Scene-level Pre-training.}
In scene pre-training, the MAE-based methods that dominate object pre-training are less effective due to the increased complexity of scenes and the limitations of Chamfer Distance supervision for reconstructed scene samples. Conversely, contrastive learning~\cite{chen2020simclr, he2020moco, long2023pointclustering, feng2024shape2scene} is particularly well-suited to capturing intricate geometric structures in scene data, pioneered by PointContrast~\cite{xie2020pc}. Subsequent methods~\cite{hou2021csc, wu2023msc, wang2024groupcontrast} introduce more efficient data augmentation and contrastive techniques. Departing from contrastive-based frameworks, PPT~\cite{wu2024ppt} explores the potential of multi-dataset pre-training, while Ponder~\cite{huang2023ponder, zhu2023ponderv2} introduces NeRF-based generative pre-training. Our proposed UniPre3D framework bears some resemblance to Ponder in approach, yet it is designed to be more efficient and is specifically developed to provide unified applicability for both object and scene pre-training.
\section{Approach}

\subsection{Preliminary: 3D Gaussian Splatting}
\label{sec:3dgs}

\noindent\textbf{Vanilla 3DGS.} 
3D Gaussian Splatting (3DGS)~\cite{kerbl20233dgs} is a highly efficient and differentiable neural rendering technique. It first predicts a set of Gaussian primitives from multi-view images, $G = \{g_k\}_{k=1}^{K}$, each defined by a mean $\mu_k$, covariance $\Sigma_k$, opacity $\alpha_k$, and spherical harmonics coefficients $\mathbf{S}_k$. These primitives are then used to render geometry-consistent novel views through compact anisotropic volumetric splats. The per-sample optimization process adaptively controls the density of the Gaussian primitives through clone and split operations.

\vspace{6pt}
\noindent\textbf{Generalizable 3DGS.} 
Generalizable 3D Gaussian Splatting methods~\cite{szymanowicz2024splatterimage, charatan2024pixelsplat, chen2025mvsplat} train a neural network $\mathcal{E}$ to directly predict Gaussian primitives: $(\mu_k, \Sigma_k, \alpha_k, \mathbf{S}_k)_k = \mathcal{E}(I)$, where $I$ can be a single-view or multi-view image. Generalizable 3DGS eliminates the need for sample-wise optimization of Gaussian primitive parameters. The neural network $\mathcal{E}$ is typically trained on a large-scale dataset containing diverse samples. It is trained to reason cross-view relations in 3D space and capture fine-grained local structures. 

\subsection{Overall Pipeline}
\noindent\textbf{Design Insights.} 
A unified 3D pre-training approach is expected to adapt to various point cloud scales and modify the pre-training task complexity accordingly. However, current 3D pre-training methods often prove to be either overly complex or too simplistic when applied to point clouds of differing scales. While scale disparity is primarily an issue in unified 3D learning, there is typically no significant gap between object-centric images and scene images in the 2D domain. Based on this observation, we propose using the image domain as an intermediary to reduce the scale differences in point cloud data. Additionally, generating projected images as the 3D pre-training task offers the advantage of adaptive difficulty. The information density of point clouds closely aligns with that of their corresponding projected images, ensuring that the pre-training task is appropriately challenging for both object and scene scales. 

Rendering projected images from 3D data has long been a challenging research problem, with numerous solutions emerging from the vibrant generation community. We propose utilizing 3D Gaussian Splatting (3DGS)~\cite{kerbl20233dgs} as the image rendering technique, in contrast to attention-based~\cite{vaswani2017transformer} predictions in TAP~\cite{wang2023tap} and Neural Radiance Field (NeRF)~\cite{mildenhall2021nerf} rendering like Ponder~\cite{huang2023ponder, zhu2023ponderv2}. Our design rationale is based on three key advantages: 3DGS demonstrates superior \textbf{scale adaptivity}, is \textbf{lightweight}, and offers \textbf{efficiency} compared to alternative methods.

Scale adaptivity is a primary consideration when designing a unified pre-training approach. As introduced in Section~\ref{sec:3dgs}, each Gaussian primitive possesses a \textit{covariance} attribute that determines its effective region. This allows for the image rendering of point clouds at any scale, where smaller point clouds are learned to be represented by Gaussian primitives with relatively large covariance, and vice versa. Given that object-centric images are generally simpler and contain less information than scene images, a slight blur resulting from a reduced number of Gaussian primitives and their larger covariance is acceptable. When 3DGS is applied to scene point clouds, the increased point density enhances the detail of the rendered images, facilitating the representation of complex geometrical structures.

Lightweight design is another critical requirement for the development of pre-training techniques. If the modules used solely for pre-training are too cumbersome, the knowledge and capacity gained during pre-training will primarily be stored in those auxiliary components, resulting in less exploitation of the backbone. In contrast to the heavy attention-based predictor in TAP~\cite{wang2023tap}, 3DGS requires only a lightweight head to predict Gaussian primitive attributes. 

The efficiency superiority of 3DGS over NeRF is also a crucial consideration. PonderV2~\cite{zhu2023ponderv2} renders only a subset of pixels for supervision due to the slow rendering speed of NeRF. In contrast, 3DGS allows our pre-training to achieve full supervision across the entire image, while also ensuring a pre-training pipeline that is approximately twice as fast as PonderV2 in scene-level experiments.

\begin{figure}[t]
  \centering
  \includegraphics[width=\linewidth]{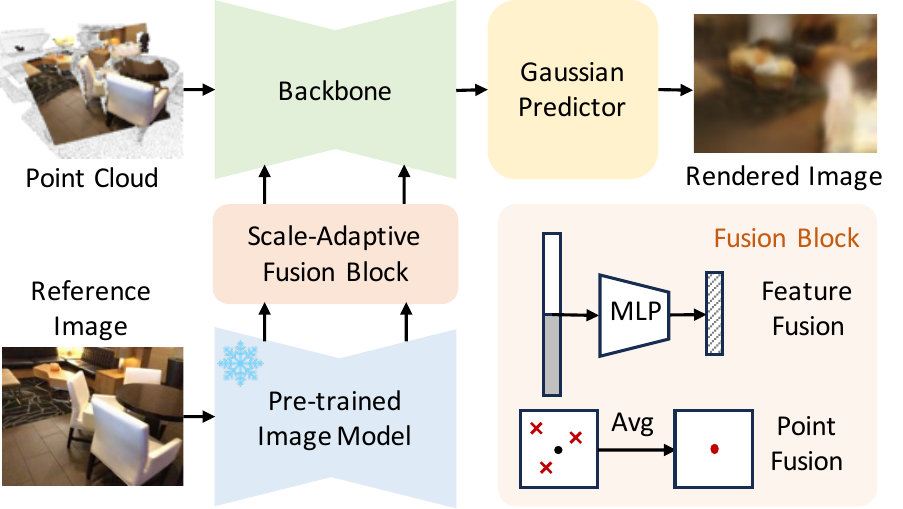}
  \vspace{-15pt}
  \caption{\textbf{UniPre3D pre-training pipeline.} Our proposed pre-training task involves predicting Gaussian parameters from the input point cloud. The 3D backbone network is expected to extract representative features, and 3D Gaussian splatting is implemented to render images for direct supervision. To incorporate additional texture information and adjust task complexity, we introduce a pre-trained image model and propose a scale-adaptive fusion block to accommodate varying data scales.}
   \label{fig:pipeline}
   \vspace{-10pt}
\end{figure}

\vspace{6pt}
\noindent\textbf{UniPre3D Pipeline.}
As outlined in the previous section, we develop a 3DGS-based framework, UniPre3D, for unified 3D pre-training, addressing the scale disparity in point cloud data. To further enhance the scale adaptability, we propose the integration of a pre-trained image model, which provides supplementary color and texture information through our novel scale-adaptive cross-modal fusion.

The overall architecture of UniPre3D illustrated in Figure~\ref{fig:pipeline} consists of two modality branches. The 3D branch includes a point cloud backbone, a lightweight Gaussian predictor, and a differentiable image renderer. The 2D branch comprises a pre-trained image feature extractor, a 2D-to-3D geometry projector, and a scale-adaptive 2D-3D fusion block. The forward propagation process is structured into three stages: \textbf{extract}, \textbf{fuse}, and \textbf{render}.

Specifically, the model accepts as input the point cloud $ P \in \mathbb{R}^{N \times 3} $ and reference projected images $ I_\mathrm{ref} \in \mathbb{R}^{V_{\mathrm{ref}} \times 3 \times H \times W} $, where $ N $ represents the number of points, $ V_{\mathrm{ref}} \geq 1 $ denotes the number of reference view images, and $ H $ and $ W $ indicate the height and width of the images, respectively. The images $ I_\mathrm{ref} $ are fed into a pre-trained image model to extract representative 2D features $ F_{\mathrm{2D}} \in \mathbb{R}^{V_{\mathrm{ref}} \times C_{\mathrm{2D}} \times H \times W} $, where $ C_{\mathrm{2D}} $ denotes the channel dimensions of the feature. These 2D features are then encoded into the 3D domain using a learnable but lightweight adaptation block $ \mathcal{A}$, followed by back-projection to the 3D space, where they are adaptively fused with the intermediate features of the point cloud model:
\begin{align}
    \hat{F}_\mathrm{2D} &= \mathcal{P}_\mathrm{2D\rightarrow 3D}(\mathcal{A}(F_\mathrm{2D})), \\
    F_\mathrm{fuse} &= \mathcal{D}(\mathcal{E}(P, \hat{F}_\mathrm{2D}), \hat{F}_\mathrm{2D}),
\end{align}
Here, $ \mathcal{E} $ and $ \mathcal{D} $ denote the encoder and decoder of the point cloud model, respectively, while $ \mathcal{P}_\mathrm{2D\rightarrow 3D} $ represents the back-projection operation from 2D to 3D. Detailed information on the back-projection process and scale-adaptive feature fusion is provided in Section~\ref{sec:fuse}.

After obtaining the fused feature $ F_\mathrm{fuse} \in \mathbb{R}^{N \times C_{\mathrm{fuse}}} $ from the point cloud backbone, we employ a lightweight Gaussian Predictor $ \mathcal{G} $ to predict Gaussian parameters $ G \in \mathbb{R}^{N \times 23} $, which encompass the Gaussian position offset, opacity, scaling, rotation quaternions, and spherical harmonics features. Finally, we utilize the differentiable Gaussian splatting technique to render $ V_\mathrm{rend} $ images $ I_r \in \mathbb{R}^{V_\mathrm{rend} \times 3 \times H \times W} $ from these Gaussian primitives.

\subsection{Scale-Adaptive Cross-Modal Fusion}
\label{sec:fuse}

\noindent\textbf{Design insights.} To modulate the difficulty of the pre-training task and enhance the point cloud model’s focus on geometry extraction, we propose the integration of pre-trained image features with the intermediate 3D features derived from the backbone model. In the context of object pre-training, the input point clouds are devoid of color, while the rendered images are expected to be rich in color. This disparity between the source and target may lead the backbone model to erroneously extract color or texture features that are irrelevant or detrimental to downstream fine-tuning. Therefore, incorporating image features from a single perspective view of the object provides essential clues regarding color and texture, while also encouraging the model to infer the color of occluded regions by fully leveraging the object’s geometry. For scene pre-training, point clouds often exhibit excessive sparsity, while the geometric relationships can be quite complex. Consequently, relying solely on point clouds as input could make the pre-training task overly challenging. Appropriately supplementing pre-trained image features can facilitate a smoother optimization process, assisting the backbone in gradually mastering the task and preventing premature convergence. Recognizing that point clouds of different scales face distinct challenges, we develop a feature fusion strategy tailored for small-scale object pre-training and propose a point fusion strategy for large-scale scene pre-training.

\vspace{6pt}
\noindent\textbf{Object-level Feature Fusion.}
Under object-level pre-training scenarios, the dataset lacks available depth maps, making the projection of 2D pixels into 3D space an ill-posed problem. Consequently, we opt to establish 2D-3D correspondence by projecting 3D points onto the 2D plane, quantizing the coordinates, and identifying the point with the minimum depth as the corresponding surface point to align with the pixel grid:
\begin{align}
    &[x, y, d, 1] = \mathrm{cat}(P,\mathbbm{1})V^{-1}, \\
    &u = \mathcal{Q}\left({xK_{[0,0]}}/{d}+K_{[0,2]}\right), \\
    &v = \mathcal{Q}\left({yK_{[1,1]}}/{d}+K_{[1,2]}\right),
\end{align}
where cat denotes concatenation, and $\mathbbm{1}$ represents an all-one tensor used to expand the dimensions of $P$ from $N \times 3$ to $N \times 4$. $\mathcal{Q}$ indicates quantization, while $K \in \mathbb{R}^{3 \times 3}$ and $V \in \mathbb{R}^{4 \times 4}$ correspond to the camera intrinsic and extrinsic matrices, respectively. The variable $d$ represents the calculated depth along the perspective projection ray for selecting the surface point, with $u \in [0, H)$ and $v \in [0, W)$ denoting pixel coordinates. For points lacking pixel correspondence, we empirically set $\hat{F}_\mathrm{2D} = 0$ to facilitate batch processing. Subsequently, we integrate the 3D feature $F_\mathrm{3D} \in \mathbb{R}^{N \times C_\mathrm{3D}}$ from the final decoder layer of the backbone with $\hat{F}_\mathrm{2D}$:
\begin{equation}
    F_\mathrm{fuse} = \mathrm{MLP}(\mathrm{cat}(F_\mathrm{3D}, \hat{F}_\mathrm{2D})),
\end{equation}
where cat refers to the concatenation operation performed along the channel dimension, while MLP denotes a multi-layer perceptron that learns to fuse cross-modal features.

\vspace{6pt}
\noindent\textbf{Scene-level Point Fusion.} In the context of scene pre-training, the aforementioned \textit{feature fusion} strategy yields unsatisfactory results, as indicated by the ablation studies in Table~\ref{tab:abl_scene}. Our analysis suggests that this is primarily due to the sparsity of the scene point cloud and the exponential increase in complexity of the generative pre-training task. To address this, we propose a \textit{point fusion} strategy for large-scale point clouds to provide enhanced visual guidance and reduce the difficulty of the pre-training task. Given that scene-level datasets include ground truth depth maps $D$, the 2D-to-3D projection can be achieved through:
\begin{align}
    [x',y',z']^T &= K^{-1}[u,v,1]^T\circ D, \\
    [x,y,z,1]^T &= V[x',y',z',1]^T,
\end{align}
Here, $(x, y, z)$ represents the back-projected pixel coordinates in 3D, $\circ$ denotes the Hadamard (element-wise) product, and $T$ indicates matrix transposition. We then treat the back-projected pixels as a pseudo point cloud $P_\mathrm{2D}$ and merge it with $P_\mathrm{3D}$, the output from the first encoding layer of the point cloud encoder, to create a cross-modal meta point cloud $P_\mathrm{meta}$. Subsequently, we perform grid sampling voxelization on $P_\mathrm{meta}$ to average the point features within each sampled voxel:
\begin{equation}
    P_\mathrm{fuse} = \mathrm{Voxelize}(\mathrm{cat}(P_\mathrm{2D}, P_\mathrm{3D})),
\end{equation}
where cat refers to the concatenation operation along the number of points dimension. The voxelized $P_\mathrm{fuse}$ is then passed through the remaining point cloud model to extract the fused features $F_\mathrm{fuse}$. Following this point fusion operation, the number of Gaussian primitives increases by 70\%, thereby enhancing both generation performance and the effectiveness of the pre-training process.

\subsection{Optimization Objectives}

We employ a pixel-wise supervision Mean Squared Error (MSE) loss during the pre-training process:
\begin{equation}
    \mathcal{L}(I_{\mathrm{r}}, I_{\mathrm{gt}}) = \frac{1}{V_\mathrm{rend}HW}\sum_{v,h,w}\left(I_{\mathrm{r}}^{v,h,w} - I_{\mathrm{gt}}^{v,h,w}\right)^2,
\end{equation}
where $I_{\mathrm{gt}}$ represents the ground truth images. Note that the $ V_\mathrm{rend} $ rendered images do not overlap with the $ V_\mathrm{ref} $ reference images in order to prevent information leakage. However, in the context of scene pre-training, we impose a restriction on the perspective gap between the reference and rendered images, maintaining it within a predefined threshold. This approach enhances the utilization of supplemental image knowledge, as a single-view image typically captures only a small portion of the scene.

In the context of object-level pre-training, we observe that object images typically follow a consistent pattern: the object is centered within the image, surrounded by background regions of solid color. To balance the pre-training difficulty between foreground and background regions, we introduce weight parameters $\omega_{\mathrm{fg}}, \omega_{\mathrm{bg}}$:
\begin{equation}
    \mathcal{L}^{\mathrm{obj}}(I_{\mathrm{r}}, I_{\mathrm{gt}}) = \omega_{\mathrm{fg}}\mathcal{L}(I^{\mathrm{fg}}_{\mathrm{r}}, I^{\mathrm{fg}}_{\mathrm{gt}}) + \omega_{\mathrm{bg}}\mathcal{L}(I^{\mathrm{bg}}_{\mathrm{r}}, I^{\mathrm{bg}}_{\mathrm{gt}}),
\end{equation}
where $I^{\mathrm{fg}}, I^{\mathrm{bg}}$ denote the foreground and background regions of each image, respectively. The boundary separating the foreground and background can be established through point cloud view projection correspondence.

\begin{figure*}[t]
  \centering
  \includegraphics[width=\linewidth]{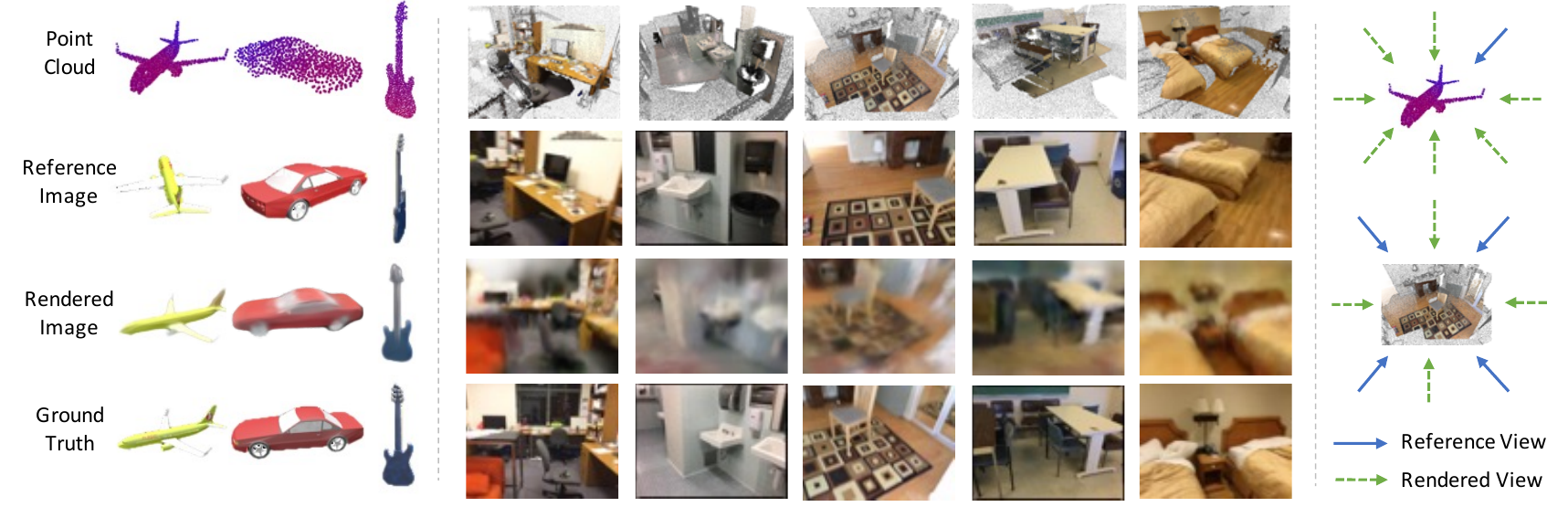}
  \vspace{-15pt}
  \caption{\textbf{Visualization of UniPre3D pre-training outputs.} The first row presents the input point clouds, followed by the reference view images in the second row. The third row displays the rendered images, which are supervised by the ground truth images shown in the fourth row. In the rightmost column, we illustrate a schematic diagram of the view selection principle for both object- and scene-level samples.}
   \label{fig:vis}
   \vspace{-10pt}
\end{figure*}

\section{Experiments}

\subsection{Pre-training}

\noindent\textbf{Data Setups.} For object-level pre-training, we adhere to established practices~\cite{pang2022pointmae, yu2022pointbert} to use the synthetic ShapeNet dataset~\cite{chang2015shapenet}. ShapeNet contains over 50,000 CAD models, from each of which we randomly sample point clouds of 1,024 points and evenly render 36 images via DISN~\cite{xu2019disn}. For scene-level pre-training, we utilize the real-world ScanNetV2 dataset~\cite{dai2017scannet} with more than 1,500 scans of indoor scenes. Each scene contains a point cloud of over 100,000 points and hundreds of associated projected images. 

\vspace{3pt}
\noindent\textbf{Backbone Model Choices.} To demonstrate the universal effectiveness of UniPre3D, we select at least a standard model and an advanced model for object- and scene-level experiments, respectively. For object-level pre-training, we begin with the standard Transformer architecture~\cite{vaswani2017transformer}, ensuring a fair comparison with previous MAE-based pre-training methods~\cite{yu2022pointbert, pang2022pointmae, liu2022maskpoint, ren2024pointcmae}. Here, \textit{standard} indicates that no structural modifications are made to the Transformer. Additionally, we pre-train on the PointMLP model~\cite{pointmlp} to enable comparison with another generative pre-training approach, TAP~\cite{wang2023tap}. To further validate our method, we pre-train on two recently proposed advanced 3D object backbone models, PointCloudMamba~\cite{zhang2024pcm} and Mamba3D~\cite{han2024mamba3d}. For scene-level pre-training, we first apply UniPre3D to the classical SparseUNet model~\cite{choy2019spunet}, allowing direct comparison with prior contrastive-based pre-training methods~\cite{xie2020pc, hou2021csc, wu2023msc}. Additionally, we use the advanced PointTransformerV3~\cite{wu2024ptv3} as the backbone, which demonstrates significantly higher baseline performance than SparseUNet, to show that UniPre3D remains effective for models with high inherent performance.

\vspace{3pt}
\noindent\textbf{Implementation Details.} 
Object models are pre-trained for 50 epochs with the Adam optimizer~\cite{kingma2014adam} and a StepLR learning rate scheduler, set to an initial learning rate of $10^{-4}$ and decaying by a factor of 0.9 every 10 epochs. The pre-training batch size is 32, with each point cloud taking one input image and supervised by four images from novel viewpoints. This requires one NVIDIA 3090Ti GPU. For scene models, we use the AdamW optimizer~\cite{loshchilov2017adamw} with a weight decay of 0.01 and an initial learning rate of $10^{-4}$. The model is pre-trained for 100 epochs and the batch size is set to 8, with each point cloud taking eight input images and supervised by eight images from novel viewpoints. We divide the ScanNet image stream into 8 bins and randomly select 1 reference view per bin. The rendered view is sampled near the reference view, with an interval restriction of fewer than 5 images. This requires eight NVIDIA 3090Ti GPUs. The pre-trained image model used as the image branch is the Stable Diffusion autoencoder~\cite{rombach2022sd}. The adaptation block $\mathcal{A}$ is implemented as a multi-layer perception (MLP) to align channel dimensions. The Gaussian predictor head is also implemented as an MLP. For object-level pre-training, the loss weight parameters are set as $\omega_{\mathrm{fg}}=4, \omega_{\mathrm{bg}}=1$.

\vspace{3pt}
\noindent\textbf{Visualization Results.} In Figure~\ref{fig:vis}, we present the rendered outputs from the pre-training stage. The first row shows part of the input point clouds, with the regions of interest highlighted for clarity. The second row displays the reference view images, followed by the rendered view images in the third row, which are supervised by the ground truth images in the fourth row. The rightmost column includes a schematic diagram illustrating the view selection principle. For object samples, only one reference view provides color cues. However, UniPre3D accurately predicts both geometry and color for other perspectives, demonstrating the 3D backbone is pre-trained to extract robust geometric features. For scene-level samples with more complex structures, although the rendered outputs are relatively blurred, the essential geometric relationships are effectively learned.

\subsection{Downstream Tasks}

\subsubsection{Object-level Fine-tuning}

\noindent\textbf{Datasets.} When fine-tuning object models for classification, we experiment on the real-world ScanObjectNN~\cite{Uy2019scanobjectnn} dataset, which comprises 15 categories and includes three splits: OBJ\_BG, OBJ\_ONLY, and PB\_T50\_RS. PB\_T50\_RS is the most challenging and significant split. For part segmentation fine-tuning, we utilize the ShapeNetPart~\cite{yi2016shapenetpart} dataset that contains over 16,000 samples across 16 classes, featuring fine-grained part annotations for 50 categories.

\begin{table}[!t]
\setlength\tabcolsep{2pt}
\caption{
\textbf{Classification results on the ScanObjectNN dataset.} We report the overall accuracy (\%) on three data splits.}
\label{tab:objcls}
\begin{center}
\vspace{-15pt}
\resizebox{1.0\linewidth}{!}{
\begin{tabular}{llccc}
    \toprule[0.95pt]
        Model & Pre-train & OBJ\_BG & OBJ\_ONLY & PB\_T50\_RS\\
    \midrule[0.6pt]
        \multirow{11}{*}{\makecell[l]{Standard \\ Transformer \\ \cite{vaswani2017transformer}}} & \xmark & 79.86 & 80.55 & 77.24 \\
        & OcCo~\cite{wang2021occo} & 84.85 & 85.54 & 78.79 \\
        & Point-BERT~\cite{yu2022pointbert} & 87.43 & 88.12 & 83.07 \\
        & MaskPoint~\cite{liu2022maskpoint} & 89.30 & 88.10 & 84.30 \\
        & Point-MAE~\cite{pang2022pointmae} & 90.02 & 88.29 & 85.18 \\
        & TAP~\cite{wang2023tap} & 90.36 & 89.50 & 85.67 \\
        & Point-CMAE~\cite{ren2024pointcmae} & 90.02 & 88.64 & 85.95 \\
        & PointDif~\cite{zheng2024pointdif} & \textbf{93.29} & 91.91 & 87.61 \\
        & \cellcolor{Gray}UniPre3D & \cellcolor{Gray}92.60  & \cellcolor{Gray}\textbf{92.08} & \cellcolor{Gray}\textbf{87.93} \\
    \midrule[0.6pt]
        \multirow{3}{*}{PointMLP~\cite{pointmlp}} & \xmark & -- & -- & 87.4 \\
        & TAP~\cite{wang2023tap} & -- & -- & 88.5 \\
        & \cellcolor{Gray}UniPre3D & \cellcolor{Gray}-- & \cellcolor{Gray}-- & \cellcolor{Gray}\textbf{89.5}\\
    \midrule[0.6pt]
        \multirow{2}{*}{PCM~\cite{zhang2024pcm}} & \xmark & -- & -- & 88.1 \\
        & \cellcolor{Gray}UniPre3D & \cellcolor{Gray}-- & \cellcolor{Gray}-- & \cellcolor{Gray}\textbf{89.0} \\
    \midrule[0.6pt]
        \multirow{2}{*}{Mamba3D~\cite{han2024mamba3d}} & \xmark & -- & -- & 92.6 \\
        & \cellcolor{Gray}UniPre3D & \cellcolor{Gray}-- & \cellcolor{Gray}-- & \cellcolor{Gray}\textbf{93.4} \\
    \bottomrule[0.95pt]
\end{tabular}}
\end{center}
\vspace{-15pt}
\end{table}

\vspace{3pt}
\noindent\textbf{Results.} For object classification in Table~\ref{tab:objcls}, UniPre3D with the standard Transformer backbone~\cite{vaswani2017transformer} outperforms others on the challenging PB\_T50\_RS benchmark. Across more advanced models~\cite{pointmlp, han2024mamba3d, zhang2024pcm}, UniPre3D delivers consistent and substantial performance gains, even on Mamba3D~\cite{han2024mamba3d} with already high accuracy. UniPre3D also surpasses the previous generative pre-training method TAP~\cite{wang2023tap} on the PointMLP~\cite{pointmlp} backbone. For part segmentation in Table~\ref{tab:partseg}, UniPre3D achieves the best performance on the mIoU$_C$ metric and competitive results with TAP on mIoU$_I$. 

\begin{table}[!t]
\setlength\tabcolsep{10pt}
\caption{\textbf{Part segmentation results on the ShapeNetPart dataset.} We report the mean IoU across all part categories mIoU$_C$, and the mean IoU across all instances mIoU$_I$.}
\label{tab:partseg}
\vspace{-5pt}
\resizebox{1.0\linewidth}{!}{
\begin{tabular}{llcc}
    \toprule[0.95pt]
        Model & Pre-train & mIoU$_C$ & mIoU$_I$\\
    \midrule[0.6pt]
    PointNet~\cite{qi2017pointnet} & \xmark & 80.4 & 83.7 \\
    PointNet++~\cite{qi2017pointnet2} & \xmark & 81.9 & 85.1 \\
    DGCNN~\cite{wang2019dgcnn} & \xmark & 82.3 & 85.2 \\
    KPConv~\cite{thomas2019kpconv} & \xmark & 85.1 & 86.4 \\
    \midrule[0.6pt]
        \multirow{7}{*}{\makecell[l]{Standard \\ Transformer \\ \cite{vaswani2017transformer}}} & \xmark & 83.4 & 84.7 \\
        & Point-BERT~\cite{yu2022pointbert} & 84.1 & 85.6 \\
        & Point-MAE~\cite{pang2022pointmae} & 84.2 & 86.1 \\
        & MaskPoint~\cite{liu2022maskpoint} & 84.4 & 86.0 \\
        & ACT~\cite{dong2022act} & 84.7 & 86.1 \\
        & Point-CMAE~\cite{ren2024pointcmae} & 84.9 & 86.0 \\
        & PCP-MAE~\cite{zhang2024pcpmae}  & 84.9 & 86.1 \\
    \midrule[0.6pt]
        \multirow{3}{*}{PointMLP~\cite{pointmlp}} & \xmark & 84.6 & 86.1 \\
        & TAP~\cite{wang2023tap} & 85.2 & \textbf{86.9} \\
        & \cellcolor{Gray}UniPre3D & \cellcolor{Gray}\textbf{85.5}  & \cellcolor{Gray}86.8 \\
    \bottomrule[0.95pt]
\end{tabular}}
\vspace{-10pt}
\end{table}

\subsubsection{Scene-level Fine-tuning}

\noindent\textbf{Datasets.} When fine-tuning on scene-level segmentation, we first assess the pre-training dataset itself, ScanNetV2~\cite{dai2017scannet}, which comprises 20 classes. Subsequently, we fine-tune on the ScanNet200~\cite{rozenberszki2022scannet200} dataset, which shares the same 2D and 3D data with ScanNetV2 but features more fine-grained annotations covering 200 categories. The classes in ScanNet200 follow a long-tail distribution, making it significantly more challenging than the ScanNetV2 dataset. We also introduce the S3DIS~\cite{armeni2016s3dis} dataset, which encompasses six areas covering 12 semantic categories.

\vspace{3pt}
\noindent\textbf{Baselines.} Point-based models~\cite{vaswani2017transformer, qian2022pointnext} pre-trained with MAE-based methods~\cite{pang2022pointmae, dong2022act, zhang2024pcpmae} on the object dataset ShapeNet~\cite{chang2015shapenet} show potential when fine-tuned on scene-level semantic segmentation. We follow their protocol to fine-tune point-based model Transformer~\cite{vaswani2017transformer} pre-trained with UniPre3D on S3DIS. However, the application of point-based models has been limited to S3DIS, and their performance still falls short of voxel-based models. Most existing scene-level pre-training methods~\cite{xie2020pc, hou2021csc, wu2023msc} rely on contrastive learning frameworks, while recent approaches~\cite{wu2024ppt, zhu2023ponderv2} have explored multi-dataset pre-training or generative pre-training with Neural Radiance Field (NeRF)~\cite{mildenhall2021nerf}. As our approach adheres to the standard protocol of single-dataset pre-training on ScanNetV2, we do not directly compare against PPT~\cite{wu2024ppt} and PonderV2~\cite{zhu2023ponderv2}, which leverage multiple datasets and supervised pre-training advantages. However, for a comprehensive evaluation, we provide reproduced results of PonderV2 under single-dataset pre-training, marked with $\dagger$ in Table~\ref{tab:semseg} and Table~\ref{tab:insseg}.

\vspace{3pt}
\noindent\textbf{Results.} For semantic segmentation in Table~\ref{tab:semseg}, UniPre3D outperforms previous object pre-training methods using the standard Transformer backbone on S3DIS. When compared to prior scene pre-training approaches with the SparseUNet backbone, UniPre3D also achieves the best results on ScanNet20 and ScanNet200. Applied to the more advanced PointTransformerV3~\cite{wu2024ptv3} backbone, UniPre3D delivers significant improvement on ScanNet200. The relatively marginal performance gain on ScanNet20 (77.45$\rightarrow$77.63) can be attributed to the near-saturation of model optimization already attained by PTv3 on this dataset. Results for PTv3 on S3DIS are omitted, as the official implementation requires disabling flash-attention, which significantly increases CUDA memory usage beyond the capacity of our NVIDIA 3090Ti GPU devices. Nonetheless, the consistent and substantial improvements delivered by UniPre3D on the more challenging ScanNet200, characterized by small objects and severe long-tail distribution, robustly demonstrate its effectiveness for scene-level pre-training. For instance segmentation in Table~\ref{tab:insseg}, UniPre3D also achieves state-of-the-art performance across most benchmarks, with particularly strong results on ScanNet200.

\begin{table}[!t]
\setlength\tabcolsep{2pt}
\caption{
\textbf{Semantic segmentation results on the scene-level datasets.} We report the mean IoU on the validation set. The Standard Transformer results are from PCP-MAE~\cite{zhang2024pcpmae}, while the voxel-based model results are from PonderV2~\cite{zhu2023ponderv2}. PPT and PonderV2 are present as grey lines only for reference, as they utilize multiple pre-training datasets or supervised pre-training, whereas we use a single dataset for unsupervised pre-training.} 
\label{tab:semseg}
\begin{center}
\vspace{-15pt}
\resizebox{1.0\linewidth}{!}{
\begin{tabular}{llccc}
    \toprule[0.95pt]
        Model & Pre-train & ScanNet20 & ScanNet200 & S3DIS\\
    \midrule[0.6pt]
        \multicolumn{5}{c}{\textit{Point-based Model}} \\
    \midrule[0.6pt]
        PointNet~\cite{qi2017pointnet} & \xmark & -- & -- & 41.1 \\
        PointNet++~\cite{qi2017pointnet2} & \xmark & -- & -- & 53.5 \\
        PointNeXt~\cite{qian2022pointnext} & \xmark & 71.5 & -- & 70.5 \\
    \midrule[0.6pt]
        \multirow{5}{*}{\makecell[l]{Standard \\ Transformer \\ \cite{vaswani2017transformer}}} & \xmark & -- & -- & 60.0 \\
        & Point-MAE~\cite{pang2022pointmae} & -- & -- & 60.8 \\
        & ACT~\cite{dong2022act} & -- & -- & 61.2 \\
        & PCP-MAE~\cite{zhang2024pcpmae} & -- & -- & 61.3 \\
        & \cellcolor{Gray}UniPre3D & \cellcolor{Gray}-- & \cellcolor{Gray}-- & \cellcolor{Gray}\textbf{62.0} \\
    \midrule[0.6pt]
        \multicolumn{5}{c}{\textit{Voxel-based Model}} \\
    \midrule[0.6pt]
        PTv1~\cite{zhao2021pt} & \xmark & 70.6 & 27.8 & 70.4 \\
        PTv2~\cite{wu2022ptv2} & \xmark & 75.4 & 30.2 & 71.6 \\
        ST~\cite{lai2022st} & \xmark & 74.3 & -- & 72.0 \\
        OctFormer~\cite{wang2023octformer} & \xmark & 75.7 & 32.6 & -- \\
    \midrule[0.6pt]
        \multirow{10}{*}{\makecell[l]{SparseUNet \\ \cite{choy2019spunet}}} & \xmark & 72.2 & 25.0 & 65.4 \\
        & PonderV1~\cite{huang2023ponder} & 73.5 & -- & -- \\
        & PC~\cite{xie2020pc} & 74.1 & 26.2 & 70.3 \\
        & CSC~\cite{hou2021csc} & 73.8 & 26.4 & \textbf{72.2} \\
        & MSC~\cite{wu2023msc} & 75.5 & 28.8 & 70.1 \\
        & \textcolor{Gray7}{PPT(Unsup.)~\cite{wu2024ppt}} & \textcolor{Gray7}{75.8} & \textcolor{Gray7}{30.4} & \textcolor{Gray7}{71.9} \\
        & \textcolor{Gray7}{PonderV2~\cite{zhu2023ponderv2}} & \textcolor{Gray7}{77.0} & \textcolor{Gray7}{32.3} & \textcolor{Gray7}{73.2} \\
        & PonderV2$^\dagger$~\cite{zhu2023ponderv2} & 74.6 & 32.4 & 70.2 \\
        & \cellcolor{Gray}UniPre3D & \cellcolor{Gray}\textbf{75.8} & \cellcolor{Gray}\textbf{33.0} & \cellcolor{Gray}71.5 \\
    \midrule[0.6pt]
        \multirow{2}{*}{PTv3~\cite{wu2024ptv3}} & \xmark & 77.5 & 35.2 & 73.4 \\
        & \cellcolor{Gray}UniPre3D & \cellcolor{Gray}\textbf{77.6} & \cellcolor{Gray}\textbf{36.0} & \cellcolor{Gray}-- \\
    \bottomrule[0.95pt]
\end{tabular}}
\end{center}
\vspace{-15pt}
\end{table}

\begin{table}[!t]
\setlength\tabcolsep{2pt}
\caption{
\textbf{Instance segmentation results on the scene-level datasets.} We use PointGroup~\cite{jiang2020pointgroup} as the baseline model, following previous papers. We report the mean average precision. PPT and PonderV2 are present as grey lines only for reference.}
\label{tab:insseg}
\begin{center}
\vspace{-15pt}
\resizebox{1.0\linewidth}{!}{
\begin{tabular}{lcccccc}
    \toprule[0.95pt]
        \multirow{2.5}{*}{Pre-train} & \multicolumn{3}{c}{ScanNet20} & \multicolumn{3}{c}{ScanNet200} \\
        \cmidrule(lr){2-4}\cmidrule(lr){5-7}
        & mAP@25 & mAP@50 & mAP & mAP@25 & mAP@50 & mAP \\ 
    \midrule[0.6pt]
        \xmark & 72.8 & 56.9 & 36.0 & 32.2 & 24.5 & 15.8 \\
        PC~\cite{xie2020pc} & -- & 58.0 & -- & -- & 24.9 & -- \\
        CSC~\cite{hou2021csc} & -- & 59.4 & -- & -- & 25.2 & -- \\
        LGround~\cite{rozenberszki2022lground} & -- & -- & -- & -- & 26.1 & -- \\
        MSC~\cite{wu2023msc} & 74.7 & 59.6 & 39.3 & 34.3 & 26.8 & 17.3 \\
        \textcolor{Gray7}{PPT (f.t.)~\cite{wu2024ppt}} & \textcolor{Gray7}{76.9} & \textcolor{Gray7}{62.0} & \textcolor{Gray7}{40.7} & \textcolor{Gray7}{36.8} & \textcolor{Gray7}{29.4} & \textcolor{Gray7}{19.4} \\
        \textcolor{Gray7}{PonderV2~\cite{zhu2023ponderv2}} & \textcolor{Gray7}{77.0} & \textcolor{Gray7}{62.6} & \textcolor{Gray7}{40.9} & \textcolor{Gray7}{37.6} & \textcolor{Gray7}{30.5} & \textcolor{Gray7}{20.1} \\
        PonderV2$^\dagger$~\cite{zhu2023ponderv2} & 75.7 & \textbf{61.7} & 39.8 & 36.0 & 28.3&18.4 \\
        \cellcolor{Gray}UniPre3D & \cellcolor{Gray}\textbf{75.9} & \cellcolor{Gray}61.3 & \cellcolor{Gray}\textbf{39.9} & \cellcolor{Gray}\textbf{37.1} & \cellcolor{Gray}\textbf{29.2} & \cellcolor{Gray}\textbf{18.7} \\
    \bottomrule[0.95pt]
\end{tabular}}
\end{center}
\vspace{-15pt}
\end{table}

\subsection{Ablation Studies}

\noindent\textbf{Integration Layer.} For object-level pre-training, we ablate on the integration layer with classification fine-tuning on ScanObjectNN (PB\_T50\_RS), shown in Table~\ref{tab:abl_obj}. For each backbone, the first row presents its baseline results, while the second row indicates pre-training with only the 3D branch. From the third to fifth rows, we progressively ablate on the integration layer used for 2D feature fusion. \textit{Decoder-Last} denotes fusion only at the final decoder layer. \textit{Decoder-Mid} represents fusion at the last two decoder layers. \textit{Decoder-All} indicates that all decoder layers are fused with image features from their corresponding layers in the image decoder. The results convey that incorporating pre-trained knowledge from the image model is crucial for improving pre-training effectiveness. However, incorporating excessive additional information may hinder fine-tuning, despite higher pre-training performance. This may stem from the model’s over-reliance on 2D features, limiting the 3D backbone to fully develop its feature extraction capacity.

\vspace{3pt}
\noindent\textbf{Fusion Strategy.} For scene-level pre-training, we ablate on the fusion strategy with semantic segmentation fine-tuning. As outlined in Section~\ref{sec:fuse}, we propose a point fusion strategy for scene pre-training to accommodate large-scale data. In Table~\ref{tab:abl_scene}, we first present pre-training with the feature fusion strategy as object pre-training in the third row. The fourth and fifth rows examine the implementation layer options for the point fusion strategy, where \textit{Enc} denotes fusion after the first layer of the backbone encoder, and \textit{Dec} denotes fusion before the final layer of the backbone decoder. The ablation results confirm our findings from object pre-training, that supplementary image knowledge is essential for enhancing our pre-training pipeline, particularly on the challenging long-tail ScanNet200 dataset. Furthermore, point fusion proves to be more effective for scene pre-training than feature fusion, with optimal fine-tuning results across all datasets achieved when fusing 2D back-projected points at the encoder layer.

\subsection{Limitations}

Even though we make an effective effort towards unified pre-training, there are still some limitations to be resolved in future research. We do not address scenarios beyond object and scene scales, and the manual fusion strategy selection further limits unification. Additionally, the requirement for both point clouds and images adds data curation burden compared with other point-only pre-training methods.

\begin{table}[!t]
\setlength\tabcolsep{8pt}
\caption{
\textbf{Ablation studies on integration layer of cross-modal feature fusion.} We report the PSNR metric for the pre-training stage and overall accuracy for the object-level fine-tuning stage.}
\label{tab:abl_obj}
\begin{center}
\vspace{-15pt}
\resizebox{1.0\linewidth}{!}{
\begin{tabular}{lcccc}
    \toprule[0.95pt]
        Model & Pre-train & Fusion & PSNR & PB\_T50\_RS \\ 
    \midrule[0.6pt]
        \multirow{4}{*}{\makecell[l]{Standard \\ Transformer \\ \cite{vaswani2017transformer}}}
        & \xmark & -- & -- & 77.2 \\
        & \cmark & \xmark & 22.8 & 86.6 \\
        & \cellcolor{Gray}\cmark & \cellcolor{Gray}Decoder-Last & 
        \cellcolor{Gray}\textbf{25.5} & \cellcolor{Gray}\textbf{87.9} \\
        & \cmark & Decoder-Mid &   23.8   & 87.0  \\
        & \cmark & Decoder-All & 24.8 & 86.5\\
    \midrule[0.6pt]
        \multirow{4}{*}{PointMLP~\cite{pointmlp}}
        & \xmark & -- & -- & 87.4 \\
        & \cmark & \xmark & 22.8 & 89.2 \\
        & \cellcolor{Gray}\cmark & \cellcolor{Gray}Decoder-Last & \cellcolor{Gray}23.8 & \cellcolor{Gray}\textbf{89.5} \\
        & \cmark & Decoder-Mid &  23.5    &    89.3 \\
        & \cmark & Decoder-All & \textbf{24.7} & 88.9 \\
    \midrule[0.6pt]
        \multirow{4}{*}{PCM~\cite{zhang2024pcm}}
        & \xmark & -- & -- & 88.1 \\
        & \cmark & \xmark & 22.5 & 88.7 \\
        & \cellcolor{Gray}\cmark & \cellcolor{Gray}Decoder-Last & \cellcolor{Gray}\textbf{23.6} & \cellcolor{Gray}\textbf{89.0} \\
        & \cmark & Decoder-Mid &  23.3    & 88.9    \\
        & \cmark & Decoder-All & 23.4 & 89.0\\
    \midrule[0.6pt]
        \multirow{4}{*}{Mamba3D~\cite{han2024mamba3d}}
        & \xmark & -- & -- & 92.6 \\
        & \cmark & \xmark & 19.8 & 93.1 \\
        & \cellcolor{Gray}\cmark & \cellcolor{Gray}Decoder-Last & \cellcolor{Gray}20.0 & \cellcolor{Gray}\textbf{93.4} \\
        & \cmark & Decoder-Mid &  20.0    &    92.9 \\
        & \cmark & Decoder-All & \textbf{20.2} & 93.1\\
    \bottomrule[0.95pt]
\end{tabular}}
\end{center}
\vspace{-10pt}
\end{table}

\begin{table}[!t]
\setlength\tabcolsep{5pt}
\caption{
\textbf{Ablation studies on cross-modal feature fusion strategies.} We report the PSNR metric for the pre-training stage and the mean IoU for the scene-level fine-tuning stage. The pre-trained model is set as the SparseUNet~\cite{choy2019spunet}.}
\label{tab:abl_scene}
\begin{center}
\vspace{-15pt}
\resizebox{1.0\linewidth}{!}{
\begin{tabular}{ccccccc}
    \toprule[0.95pt]
        Pre-train & Fusion & Layer & PSNR & ScanNet20 & ScanNet200 & S3DIS \\ 
    \midrule[0.6pt]
        \xmark & -- & -- & -- & 72.2 & 25.0 & 65.4 \\
        \cmark & \xmark & -- & 16.6 & 75.5 & 30.8 & 71.1 \\
        \cmark & Feat & Dec & 16.7 &75.7 &32.3& 70.9 \\
        \cmark & Point & Dec & 16.6 &75.7 &32.8& 70.8 \\
        \cellcolor{Gray}\cmark & \cellcolor{Gray}Point & \cellcolor{Gray}Enc & \cellcolor{Gray}\textbf{16.8} & \cellcolor{Gray}\textbf{75.8} & \cellcolor{Gray}\textbf{33.0} & \cellcolor{Gray}\textbf{71.5}  \\ 
    \bottomrule[0.95pt]
\end{tabular}}
\end{center}
\vspace{-15pt}
\end{table}

\section{Conclusion}

In this paper, we propose UniPre3D, a unified pre-training framework that is effective across point clouds of various scales. The designed pre-training task renders view images from point clouds via the efficient and differentiable 3D Gaussian splatting. To adaptively control task complexity and enable the backbone model to prioritize geometry feature extraction, we propose scale-adaptive fusion techniques that integrate pre-trained image features with 3D features. For object pre-training, we implement feature fusion to provide texture and color cues, while for scene pre-training, we propose point fusion to densify sparse scene point clouds and provide visual support. Our extensive experiments on standard and advanced point cloud models across both object and scene perception tasks demonstrate the universal effectiveness of UniPre3D. Our unified approach consistently outperforms prior scale-specific pre-training methods on most benchmarks, underscoring its robustness and adaptability. Furthermore, we conduct thorough ablations to discuss knowledge integration layer choices and multiple fusion strategies. We believe this work will inspire future research on unified model architectures and pre-training strategies within the 3D domain.

\section*{Acknowledgments}
This work was supported in part by the National Natural Science Foundation of China under Grant 623B2063, Grant 62321005, Grant 62336004, and Grant 62125603.

{
    \small
    \bibliographystyle{ieeenat_fullname}
    \bibliography{main}

\begin{thebibliography}{85}
\providecommand{\natexlab}[1]{#1}
\providecommand{\url}[1]{\texttt{#1}}
\expandafter\ifx\csname urlstyle\endcsname\relax
  \providecommand{\doi}[1]{doi: #1}\else
  \providecommand{\doi}{doi: \begingroup \urlstyle{rm}\Url}\fi

\bibitem[Armeni et~al.(2016)Armeni, Sener, Zamir, Jiang, Brilakis, Fischer, and Savarese]{armeni2016s3dis}
Iro Armeni, Ozan Sener, Amir~R Zamir, Helen Jiang, Ioannis Brilakis, Martin Fischer, and Silvio Savarese.
\newblock 3d semantic parsing of large-scale indoor spaces.
\newblock In \emph{CVPR}, 2016.

\bibitem[Chang et~al.(2015)Chang, Funkhouser, Guibas, Hanrahan, Huang, Li, Savarese, Savva, Song, Su, et~al.]{chang2015shapenet}
Angel~X Chang, Thomas Funkhouser, Leonidas Guibas, Pat Hanrahan, Qixing Huang, Zimo Li, Silvio Savarese, Manolis Savva, Shuran Song, Hao Su, et~al.
\newblock Shapenet: An information-rich 3d model repository.
\newblock \emph{arXiv preprint arXiv:1512.03012}, 2015.

\bibitem[Charatan et~al.(2024)Charatan, Li, Tagliasacchi, and Sitzmann]{charatan2024pixelsplat}
David Charatan, Sizhe~Lester Li, Andrea Tagliasacchi, and Vincent Sitzmann.
\newblock pixelsplat: 3d gaussian splats from image pairs for scalable generalizable 3d reconstruction.
\newblock In \emph{CVPR}, 2024.

\bibitem[Chen et~al.(2023{\natexlab{a}})Chen, Xia, Zang, Wang, and Li]{chen2023dela}
Binjie Chen, Yunzhou Xia, Yu Zang, Cheng Wang, and Jonathan Li.
\newblock Decoupled local aggregation for point cloud learning.
\newblock \emph{arXiv preprint arXiv:2308.16532}, 2023{\natexlab{a}}.

\bibitem[Chen et~al.(2024)Chen, Wang, Yang, Yu, Yuan, and Yue]{chen2024pointgpt}
Guangyan Chen, Meiling Wang, Yi Yang, Kai Yu, Li Yuan, and Yufeng Yue.
\newblock Pointgpt: Auto-regressively generative pre-training from point clouds.
\newblock \emph{NeurIPS}, 2024.

\bibitem[Chen et~al.(2020)Chen, Kornblith, Norouzi, and Hinton]{chen2020simclr}
Ting Chen, Simon Kornblith, Mohammad Norouzi, and Geoffrey Hinton.
\newblock A simple framework for contrastive learning of visual representations.
\newblock In \emph{ICML}, 2020.

\bibitem[Chen et~al.(2025)Chen, Xu, Zheng, Zhuang, Pollefeys, Geiger, Cham, and Cai]{chen2025mvsplat}
Yuedong Chen, Haofei Xu, Chuanxia Zheng, Bohan Zhuang, Marc Pollefeys, Andreas Geiger, Tat-Jen Cham, and Jianfei Cai.
\newblock Mvsplat: Efficient 3d gaussian splatting from sparse multi-view images.
\newblock In \emph{ECCV}, 2025.

\bibitem[Chen et~al.(2023{\natexlab{b}})Chen, Hu, Chen, Nie{\ss}ner, and Chang]{chen2023unit3d}
Zhenyu Chen, Ronghang Hu, Xinlei Chen, Matthias Nie{\ss}ner, and Angel~X Chang.
\newblock Unit3d: A unified transformer for 3d dense captioning and visual grounding.
\newblock In \emph{ICCV}, 2023{\natexlab{b}}.

\bibitem[Choy et~al.(2019)Choy, Gwak, and Savarese]{choy2019spunet}
Christopher Choy, JunYoung Gwak, and Silvio Savarese.
\newblock 4d spatio-temporal convnets: Minkowski convolutional neural networks.
\newblock In \emph{CVPR}, 2019.

\bibitem[Contributors(2020)]{mmdet3d2020}
MMDetection3D Contributors.
\newblock {MMDetection3D: OpenMMLab} next-generation platform for general {3D} object detection.
\newblock \url{https://github.com/open-mmlab/mmdetection3d}, 2020.

\bibitem[Dai et~al.(2017)Dai, Chang, Savva, Halber, Funkhouser, and Nie{\ss}ner]{dai2017scannet}
Angela Dai, Angel~X Chang, Manolis Savva, Maciej Halber, Thomas Funkhouser, and Matthias Nie{\ss}ner.
\newblock Scannet: Richly-annotated 3d reconstructions of indoor scenes.
\newblock In \emph{CVPR}, 2017.

\bibitem[Dong et~al.(2022)Dong, Qi, Zhang, Zhang, Sun, Ge, Yi, and Ma]{dong2022act}
Runpei Dong, Zekun Qi, Linfeng Zhang, Junbo Zhang, Jianjian Sun, Zheng Ge, Li Yi, and Kaisheng Ma.
\newblock Autoencoders as cross-modal teachers: Can pretrained 2d image transformers help 3d representation learning?
\newblock \emph{arXiv preprint arXiv:2212.08320}, 2022.

\bibitem[Feng et~al.(2024{\natexlab{a}})Feng, Quan, Wang, Wang, and Yang]{feng2024interpretable3d}
Tuo Feng, Ruijie Quan, Xiaohan Wang, Wenguan Wang, and Yi Yang.
\newblock Interpretable3d: An ad-hoc interpretable classifier for 3d point clouds.
\newblock In \emph{NeurIPS}, 2024{\natexlab{a}}.

\bibitem[Feng et~al.(2024{\natexlab{b}})Feng, Wang, Quan, and Yang]{feng2024shape2scene}
Tuo Feng, Wenguan Wang, Ruijie Quan, and Yi Yang.
\newblock Shape2scene: 3d scene representation learning through pre-training on shape data.
\newblock In \emph{ECCV}, 2024{\natexlab{b}}.

\bibitem[Han et~al.(2024)Han, Tang, Wang, and Li]{han2024mamba3d}
Xu Han, Yuan Tang, Zhaoxuan Wang, and Xianzhi Li.
\newblock Mamba3d: Enhancing local features for 3d point cloud analysis via state space model.
\newblock \emph{arXiv preprint arXiv:2404.14966}, 2024.

\bibitem[He et~al.(2020)He, Fan, Wu, Xie, and Girshick]{he2020moco}
Kaiming He, Haoqi Fan, Yuxin Wu, Saining Xie, and Ross Girshick.
\newblock Momentum contrast for unsupervised visual representation learning.
\newblock In \emph{CVPR}, 2020.

\bibitem[He et~al.(2022)He, Chen, Xie, Li, Doll{\'a}r, and Girshick]{he2022mae}
Kaiming He, Xinlei Chen, Saining Xie, Yanghao Li, Piotr Doll{\'a}r, and Ross Girshick.
\newblock Masked autoencoders are scalable vision learners.
\newblock In \emph{CVPR}, 2022.

\bibitem[Hou et~al.(2021)Hou, Graham, Nie{\ss}ner, and Xie]{hou2021csc}
Ji Hou, Benjamin Graham, Matthias Nie{\ss}ner, and Saining Xie.
\newblock Exploring data-efficient 3d scene understanding with contrastive scene contexts.
\newblock In \emph{CVPR}, 2021.

\bibitem[Huang et~al.(2023)Huang, Peng, He, Yang, Zhou, and Ouyang]{huang2023ponder}
Di Huang, Sida Peng, Tong He, Honghui Yang, Xiaowei Zhou, and Wanli Ouyang.
\newblock Ponder: Point cloud pre-training via neural rendering.
\newblock In \emph{ICCV}, 2023.

\bibitem[Huang et~al.(2021)Huang, Xie, Zhu, and Zhu]{huang2021strl}
Siyuan Huang, Yichen Xie, Song-Chun Zhu, and Yixin Zhu.
\newblock Spatio-temporal self-supervised representation learning for 3d point clouds.
\newblock In \emph{ICCV}, 2021.

\bibitem[Jiang et~al.(2020)Jiang, Zhao, Shi, Liu, Fu, and Jia]{jiang2020pointgroup}
Li Jiang, Hengshuang Zhao, Shaoshuai Shi, Shu Liu, Chi-Wing Fu, and Jiaya Jia.
\newblock Pointgroup: Dual-set point grouping for 3d instance segmentation.
\newblock In \emph{CVPR}, 2020.

\bibitem[Kerbl et~al.(2023)Kerbl, Kopanas, Leimk{\"u}hler, and Drettakis]{kerbl20233dgs}
Bernhard Kerbl, Georgios Kopanas, Thomas Leimk{\"u}hler, and George Drettakis.
\newblock 3d gaussian splatting for real-time radiance field rendering.
\newblock \emph{ACM Trans. Graph.}, 2023.

\bibitem[Kingma(2014)]{kingma2014adam}
Diederik~P Kingma.
\newblock Adam: A method for stochastic optimization.
\newblock \emph{arXiv preprint arXiv:1412.6980}, 2014.

\bibitem[Kolodiazhnyi et~al.(2024)Kolodiazhnyi, Vorontsova, Konushin, and Rukhovich]{kolodiazhnyi2024oneformer3d}
Maxim Kolodiazhnyi, Anna Vorontsova, Anton Konushin, and Danila Rukhovich.
\newblock Oneformer3d: One transformer for unified point cloud segmentation.
\newblock In \emph{CVPR}, 2024.

\bibitem[Lai et~al.(2022)Lai, Liu, Jiang, Wang, Zhao, Liu, Qi, and Jia]{lai2022st}
Xin Lai, Jianhui Liu, Li Jiang, Liwei Wang, Hengshuang Zhao, Shu Liu, Xiaojuan Qi, and Jiaya Jia.
\newblock Stratified transformer for 3d point cloud segmentation.
\newblock In \emph{CVPR}, 2022.

\bibitem[Liu et~al.(2022)Liu, Cai, and Lee]{liu2022maskpoint}
Haotian Liu, Mu Cai, and Yong~Jae Lee.
\newblock Masked discrimination for self-supervised learning on point clouds.
\newblock In \emph{ECCV}, 2022.

\bibitem[Liu et~al.(2023)Liu, Chen, Wang, King, and Liu]{liu2023pointrae}
Yang Liu, Chen Chen, Can Wang, Xulin King, and Mengyuan Liu.
\newblock Regress before construct: Regress autoencoder for point cloud self-supervised learning.
\newblock In \emph{ACM MM}, 2023.

\bibitem[Long et~al.(2023)Long, Yao, Qiu, Li, and Mei]{long2023pointclustering}
Fuchen Long, Ting Yao, Zhaofan Qiu, Lusong Li, and Tao Mei.
\newblock Pointclustering: Unsupervised point cloud pre-training using transformation invariance in clustering.
\newblock In \emph{CVPR}, 2023.

\bibitem[Loshchilov(2017)]{loshchilov2017adamw}
I Loshchilov.
\newblock Decoupled weight decay regularization.
\newblock \emph{arXiv preprint arXiv:1711.05101}, 2017.

\bibitem[Lu et~al.(2022)Lu, Clark, Zellers, Mottaghi, and Kembhavi]{lu2022unifiedio}
Jiasen Lu, Christopher Clark, Rowan Zellers, Roozbeh Mottaghi, and Aniruddha Kembhavi.
\newblock Unified-io: A unified model for vision, language, and multi-modal tasks.
\newblock In \emph{ICLR}, 2022.

\bibitem[Ma et~al.(2022)Ma, Qin, You, Ran, and Fu]{pointmlp}
Xu Ma, Can Qin, Haoxuan You, Haoxi Ran, and Yun Fu.
\newblock Rethinking network design and local geometry in point cloud: A simple residual mlp framework.
\newblock In \emph{ICLR}, 2022.

\bibitem[Mildenhall et~al.(2021)Mildenhall, Srinivasan, Tancik, Barron, Ramamoorthi, and Ng]{mildenhall2021nerf}
Ben Mildenhall, Pratul~P Srinivasan, Matthew Tancik, Jonathan~T Barron, Ravi Ramamoorthi, and Ren Ng.
\newblock Nerf: Representing scenes as neural radiance fields for view synthesis.
\newblock \emph{Communications of the ACM}, 2021.

\bibitem[Pang et~al.(2022)Pang, Wang, Tay, Liu, Tian, and Yuan]{pang2022pointmae}
Yatian Pang, Wenxiao Wang, Francis~EH Tay, Wei Liu, Yonghong Tian, and Li Yuan.
\newblock Masked autoencoders for point cloud self-supervised learning.
\newblock In \emph{ECCV}, 2022.

\bibitem[Park et~al.(2023)Park, Lee, Kim, Xiong, and Kim]{park2023spotr}
Jinyoung Park, Sanghyeok Lee, Sihyeon Kim, Yunyang Xiong, and Hyunwoo~J Kim.
\newblock Self-positioning point-based transformer for point cloud understanding.
\newblock In \emph{CVPR}, 2023.

\bibitem[Peng et~al.(2024)Peng, Wu, Jiang, Chen, Zhao, Tian, and Jia]{peng2024oacnn}
Bohao Peng, Xiaoyang Wu, Li Jiang, Yukang Chen, Hengshuang Zhao, Zhuotao Tian, and Jiaya Jia.
\newblock Oa-cnns: Omni-adaptive sparse cnns for 3d semantic segmentation.
\newblock In \emph{CVPR}, 2024.

\bibitem[Qi et~al.(2017{\natexlab{a}})Qi, Su, Mo, and Guibas]{qi2017pointnet}
Charles~R Qi, Hao Su, Kaichun Mo, and Leonidas~J Guibas.
\newblock Pointnet: Deep learning on point sets for 3d classification and segmentation.
\newblock In \emph{CVPR}, 2017{\natexlab{a}}.

\bibitem[Qi et~al.(2017{\natexlab{b}})Qi, Yi, Su, and Guibas]{qi2017pointnet2}
Charles~R Qi, Li Yi, Hao Su, and Leonidas~J Guibas.
\newblock Pointnet++ deep hierarchical feature learning on point sets in a metric space.
\newblock In \emph{NeurIPS}, 2017{\natexlab{b}}.

\bibitem[Qi et~al.(2019)Qi, Litany, He, and Guibas]{qi2019votenet}
Charles~R Qi, Or Litany, Kaiming He, and Leonidas~J Guibas.
\newblock Deep hough voting for 3d object detection in point clouds.
\newblock In \emph{ICCV}, 2019.

\bibitem[Qi et~al.(2023)Qi, Dong, Fan, Ge, Zhang, Ma, and Yi]{qi2023recon}
Zekun Qi, Runpei Dong, Guofan Fan, Zheng Ge, Xiangyu Zhang, Kaisheng Ma, and Li Yi.
\newblock Contrast with reconstruct: Contrastive 3d representation learning guided by generative pretraining.
\newblock \emph{arXiv preprint arXiv:2302.02318}, 2023.

\bibitem[Qi et~al.(2024)Qi, Yu, Dong, and Ma]{qi2024vpp}
Zekun Qi, Muzhou Yu, Runpei Dong, and Kaisheng Ma.
\newblock Vpp: Efficient conditional 3d generation via voxel-point progressive representation.
\newblock \emph{NeurIPS}, 2024.

\bibitem[Qi et~al.(2025)Qi, Dong, Zhang, Geng, Han, Ge, Yi, and Ma]{qi2025recon++}
Zekun Qi, Runpei Dong, Shaochen Zhang, Haoran Geng, Chunrui Han, Zheng Ge, Li Yi, and Kaisheng Ma.
\newblock Shapellm: Universal 3d object understanding for embodied interaction.
\newblock In \emph{ECCV}, 2025.

\bibitem[Qian et~al.(2022)Qian, Li, Peng, Mai, Hammoud, Elhoseiny, and Ghanem]{qian2022pointnext}
Guocheng Qian, Yuchen Li, Houwen Peng, Jinjie Mai, Hasan Hammoud, Mohamed Elhoseiny, and Bernard Ghanem.
\newblock Pointnext: Revisiting pointnet++ with improved training and scaling strategies.
\newblock In \emph{NeurIPS}, 2022.

\bibitem[Rao et~al.(2021)Rao, Liu, Wei, Lu, Hsieh, and Zhou]{rao2021randomrooms}
Yongming Rao, Benlin Liu, Yi Wei, Jiwen Lu, Cho-Jui Hsieh, and Jie Zhou.
\newblock Randomrooms: Unsupervised pre-training from synthetic shapes and randomized layouts for 3d object detection.
\newblock In \emph{ICCV}, 2021.

\bibitem[Ren et~al.(2024)Ren, Mei, Paudel, Wang, Li, Liu, Cucchiara, Van~Gool, and Sebe]{ren2024pointcmae}
Bin Ren, Guofeng Mei, Danda~Pani Paudel, Weijie Wang, Yawei Li, Mengyuan Liu, Rita Cucchiara, Luc Van~Gool, and Nicu Sebe.
\newblock Bringing masked autoencoders explicit contrastive properties for point cloud self-supervised learning.
\newblock \emph{arXiv preprint arXiv:2407.05862}, 2024.

\bibitem[Rombach et~al.(2022)Rombach, Blattmann, Lorenz, Esser, and Ommer]{rombach2022sd}
Robin Rombach, Andreas Blattmann, Dominik Lorenz, Patrick Esser, and Bj{\"o}rn Ommer.
\newblock High-resolution image synthesis with latent diffusion models.
\newblock In \emph{CVPR}, 2022.

\bibitem[Rozenberszki et~al.(2022{\natexlab{a}})Rozenberszki, Litany, and Dai]{rozenberszki2022lground}
David Rozenberszki, Or Litany, and Angela Dai.
\newblock Language-grounded indoor 3d semantic segmentation in the wild.
\newblock In \emph{ECCV}, 2022{\natexlab{a}}.

\bibitem[Rozenberszki et~al.(2022{\natexlab{b}})Rozenberszki, Litany, and Dai]{rozenberszki2022scannet200}
David Rozenberszki, Or Litany, and Angela Dai.
\newblock Language-grounded indoor 3d semantic segmentation in the wild.
\newblock In \emph{ECCV}, 2022{\natexlab{b}}.

\bibitem[Szymanowicz et~al.(2024)Szymanowicz, Rupprecht, and Vedaldi]{szymanowicz2024splatterimage}
Stanislaw Szymanowicz, Chrisitian Rupprecht, and Andrea Vedaldi.
\newblock Splatter image: Ultra-fast single-view 3d reconstruction.
\newblock In \emph{CVPR}, 2024.

\bibitem[Thomas et~al.(2019)Thomas, Qi, Deschaud, Marcotegui, Goulette, and Guibas]{thomas2019kpconv}
Hugues Thomas, Charles~R. Qi, Jean-Emmanuel Deschaud, Beatriz Marcotegui, Fran{\c{c}}ois Goulette, and Leonidas~J. Guibas.
\newblock Kpconv: Flexible and deformable convolution for point clouds.
\newblock \emph{ICCV}, 2019.

\bibitem[Thomas et~al.(2024)Thomas, Tsai, Barfoot, and Zhang]{thomas2024kpconvx}
Hugues Thomas, Yao-Hung~Hubert Tsai, Timothy~D Barfoot, and Jian Zhang.
\newblock Kpconvx: Modernizing kernel point convolution with kernel attention.
\newblock In \emph{CVPR}, 2024.

\bibitem[Uy et~al.(2019)Uy, Pham, Hua, Nguyen, and Yeung]{Uy2019scanobjectnn}
Mikaela~Angelina Uy, Quang-Hieu Pham, Binh-Son Hua, Duc~Thanh Nguyen, and Sai-Kit Yeung.
\newblock Revisiting point cloud classification: A new benchmark dataset and classification model on real-world data.
\newblock In \emph{ICCV}, 2019.

\bibitem[Vaswani et~al.(2017)Vaswani, Shazeer, Parmar, Uszkoreit, Jones, Gomez, Kaiser, and Polosukhin]{vaswani2017transformer}
Ashish Vaswani, Noam Shazeer, Niki Parmar, Jakob Uszkoreit, Llion Jones, Aidan~N Gomez, {\L}ukasz Kaiser, and Illia Polosukhin.
\newblock Attention is all you need.
\newblock In \emph{NeurIPS}, 2017.

\bibitem[Wang et~al.(2024{\natexlab{a}})Wang, Jiang, Wu, Tian, Peng, Zhao, and Jia]{wang2024groupcontrast}
Chengyao Wang, Li Jiang, Xiaoyang Wu, Zhuotao Tian, Bohao Peng, Hengshuang Zhao, and Jiaya Jia.
\newblock Groupcontrast: Semantic-aware self-supervised representation learning for 3d understanding.
\newblock In \emph{CVPR}, 2024{\natexlab{a}}.

\bibitem[Wang et~al.(2024{\natexlab{b}})Wang, Wu, Lam, Ning, Yu, Wang, Li, and Srikanthan]{wang2024gpsformer}
Changshuo Wang, Meiqing Wu, Siew-Kei Lam, Xin Ning, Shangshu Yu, Ruiping Wang, Weijun Li, and Thambipillai Srikanthan.
\newblock Gpsformer: A global perception and local structure fitting-based transformer for point cloud understanding.
\newblock \emph{arXiv preprint arXiv:2407.13519}, 2024{\natexlab{b}}.

\bibitem[Wang et~al.(2021)Wang, Liu, Yue, Lasenby, and Kusner]{wang2021occo}
Hanchen Wang, Qi Liu, Xiangyu Yue, Joan Lasenby, and Matt~J Kusner.
\newblock Unsupervised point cloud pre-training via occlusion completion.
\newblock In \emph{ICCV}, 2021.

\bibitem[Wang et~al.(2023{\natexlab{a}})Wang, Tang, Ji, Sun, Zhang, Ma, Zhao, Li, Zhao, Lv, et~al.]{wang2023jm3d}
Haowei Wang, Jiji Tang, Jiayi Ji, Xiaoshuai Sun, Rongsheng Zhang, Yiwei Ma, Minda Zhao, Lincheng Li, Zeng Zhao, Tangjie Lv, et~al.
\newblock Beyond first impressions: Integrating joint multi-modal cues for comprehensive 3d representation.
\newblock In \emph{ACM MM}, pages 3403--3414, 2023{\natexlab{a}}.

\bibitem[Wang(2023)]{wang2023octformer}
Peng-Shuai Wang.
\newblock Octformer: Octree-based transformers for 3d point clouds.
\newblock \emph{TOG}, 2023.

\bibitem[Wang et~al.(2023{\natexlab{b}})Wang, Bao, Dong, Bjorck, Peng, Liu, Aggarwal, Mohammed, Singhal, Som, et~al.]{wang2023beitv3}
Wenhui Wang, Hangbo Bao, Li Dong, Johan Bjorck, Zhiliang Peng, Qiang Liu, Kriti Aggarwal, Owais~Khan Mohammed, Saksham Singhal, Subhojit Som, et~al.
\newblock Image as a foreign language: Beit pretraining for vision and vision-language tasks.
\newblock In \emph{CVPR}, 2023{\natexlab{b}}.

\bibitem[Wang et~al.(2019)Wang, Sun, Liu, Sarma, Bronstein, and Solomon]{wang2019dgcnn}
Yue Wang, Yongbin Sun, Ziwei Liu, Sanjay~E Sarma, Michael~M Bronstein, and Justin~M Solomon.
\newblock Dynamic graph cnn for learning on point clouds.
\newblock \emph{TOG}, 2019.

\bibitem[Wang et~al.(2023{\natexlab{c}})Wang, Yu, Rao, Zhou, and Lu]{wang2023tap}
Ziyi Wang, Xumin Yu, Yongming Rao, Jie Zhou, and Jiwen Lu.
\newblock Take-a-photo: 3d-to-2d generative pre-training of point cloud models.
\newblock In \emph{ICCV}, 2023{\natexlab{c}}.

\bibitem[Wu et~al.(2022)Wu, Lao, Jiang, Liu, and Zhao]{wu2022ptv2}
Xiaoyang Wu, Yixing Lao, Li Jiang, Xihui Liu, and Hengshuang Zhao.
\newblock Point transformer v2: Grouped vector attention and partition-based pooling.
\newblock In \emph{NeurIPS}, 2022.

\bibitem[Wu et~al.(2023)Wu, Wen, Liu, and Zhao]{wu2023msc}
Xiaoyang Wu, Xin Wen, Xihui Liu, and Hengshuang Zhao.
\newblock Masked scene contrast: A scalable framework for unsupervised 3d representation learning.
\newblock In \emph{CVPR}, 2023.

\bibitem[Wu et~al.(2024{\natexlab{a}})Wu, Jiang, Wang, Liu, Liu, Qiao, Ouyang, He, and Zhao]{wu2024ptv3}
Xiaoyang Wu, Li Jiang, Peng-Shuai Wang, Zhijian Liu, Xihui Liu, Yu Qiao, Wanli Ouyang, Tong He, and Hengshuang Zhao.
\newblock Point transformer v3: Simpler faster stronger.
\newblock In \emph{CVPR}, 2024{\natexlab{a}}.

\bibitem[Wu et~al.(2024{\natexlab{b}})Wu, Tian, Wen, Peng, Liu, Yu, and Zhao]{wu2024ppt}
Xiaoyang Wu, Zhuotao Tian, Xin Wen, Bohao Peng, Xihui Liu, Kaicheng Yu, and Hengshuang Zhao.
\newblock Towards large-scale 3d representation learning with multi-dataset point prompt training.
\newblock In \emph{CVPR}, 2024{\natexlab{b}}.

\bibitem[Xie et~al.(2020)Xie, Gu, Guo, Qi, Guibas, and Litany]{xie2020pc}
Saining Xie, Jiatao Gu, Demi Guo, Charles~R Qi, Leonidas Guibas, and Or Litany.
\newblock Pointcontrast: Unsupervised pre-training for 3d point cloud understanding.
\newblock In \emph{ECCV}, 2020.

\bibitem[Xu et~al.(2019)Xu, Wang, Ceylan, Mech, and Neumann]{xu2019disn}
Qiangeng Xu, Weiyue Wang, Duygu Ceylan, Radomir Mech, and Ulrich Neumann.
\newblock Disn: Deep implicit surface network for high-quality single-view 3d reconstruction.
\newblock In \emph{NeurIPS}, 2019.

\bibitem[Xue et~al.(2023)Xue, Gao, Xing, Mart{\'\i}n-Mart{\'\i}n, Wu, Xiong, Xu, Niebles, and Savarese]{xue2023ulip}
Le Xue, Mingfei Gao, Chen Xing, Roberto Mart{\'\i}n-Mart{\'\i}n, Jiajun Wu, Caiming Xiong, Ran Xu, Juan~Carlos Niebles, and Silvio Savarese.
\newblock Ulip: Learning a unified representation of language, images, and point clouds for 3d understanding.
\newblock In \emph{CVPR}, 2023.

\bibitem[Xue et~al.(2024)Xue, Yu, Zhang, Panagopoulou, Li, Mart{\'\i}n-Mart{\'\i}n, Wu, Xiong, Xu, Niebles, et~al.]{xue2024ulip2}
Le Xue, Ning Yu, Shu Zhang, Artemis Panagopoulou, Junnan Li, Roberto Mart{\'\i}n-Mart{\'\i}n, Jiajun Wu, Caiming Xiong, Ran Xu, Juan~Carlos Niebles, et~al.
\newblock Ulip-2: Towards scalable multimodal pre-training for 3d understanding.
\newblock In \emph{CVPR}, 2024.

\bibitem[Yamada et~al.(2022)Yamada, Kataoka, Chiba, Domae, and Ogata]{yamada2022pcfractaldb}
Ryosuke Yamada, Hirokatsu Kataoka, Naoya Chiba, Yukiyasu Domae, and Tetsuya Ogata.
\newblock Point cloud pre-training with natural 3d structures.
\newblock In \emph{CVPR}, 2022.

\bibitem[Yan et~al.(2023)Yan, Yang, Li, Song, Guan, Kang, Hua, and Huang]{yan2023iae}
Siming Yan, Zhenpei Yang, Haoxiang Li, Chen Song, Li Guan, Hao Kang, Gang Hua, and Qixing Huang.
\newblock Implicit autoencoder for point-cloud self-supervised representation learning.
\newblock In \emph{ICCV}, 2023.

\bibitem[Yang et~al.(2023)Yang, Guo, Xiong, Liu, Pan, Wang, Tong, and Guo]{yang2023swin3d}
Yu-Qi Yang, Yu-Xiao Guo, Jian-Yu Xiong, Yang Liu, Hao Pan, Peng-Shuai Wang, Xin Tong, and Baining Guo.
\newblock Swin3d: A pretrained transformer backbone for 3d indoor scene understanding.
\newblock \emph{arXiv preprint arXiv:2304.06906}, 2023.

\bibitem[Yi et~al.(2016)Yi, Kim, Ceylan, Shen, Yan, Su, Lu, Huang, Sheffer, and Guibas]{yi2016shapenetpart}
Li Yi, Vladimir~G Kim, Duygu Ceylan, I-Chao Shen, Mengyan Yan, Hao Su, Cewu Lu, Qixing Huang, Alla Sheffer, and Leonidas Guibas.
\newblock A scalable active framework for region annotation in 3d shape collections.
\newblock \emph{ToG}, 2016.

\bibitem[Yu et~al.(2022)Yu, Tang, Rao, Huang, Zhou, and Lu]{yu2022pointbert}
Xumin Yu, Lulu Tang, Yongming Rao, Tiejun Huang, Jie Zhou, and Jiwen Lu.
\newblock Point-bert: Pre-training 3d point cloud transformers with masked point modeling.
\newblock In \emph{CVPR}, 2022.

\bibitem[Zha et~al.(2024)Zha, Ji, Li, Li, Dai, Chen, Wang, and Xia]{zha2024pointfemae}
Yaohua Zha, Huizhen Ji, Jinmin Li, Rongsheng Li, Tao Dai, Bin Chen, Zhi Wang, and Shu-Tao Xia.
\newblock Towards compact 3d representations via point feature enhancement masked autoencoders.
\newblock In \emph{AAAI}, 2024.

\bibitem[Zhang et~al.(2022)Zhang, Guo, Gao, Fang, Zhao, Wang, Qiao, and Li]{zhang2022pointm2ae}
Renrui Zhang, Ziyu Guo, Peng Gao, Rongyao Fang, Bin Zhao, Dong Wang, Yu Qiao, and Hongsheng Li.
\newblock Point-m2ae: multi-scale masked autoencoders for hierarchical point cloud pre-training.
\newblock \emph{NeurIPS}, 2022.

\bibitem[Zhang et~al.(2023{\natexlab{a}})Zhang, Wang, Qiao, Gao, and Li]{zhang2023i2pmae}
Renrui Zhang, Liuhui Wang, Yu Qiao, Peng Gao, and Hongsheng Li.
\newblock Learning 3d representations from 2d pre-trained models via image-to-point masked autoencoders.
\newblock In \emph{CVPR}, 2023{\natexlab{a}}.

\bibitem[Zhang et~al.(2024{\natexlab{a}})Zhang, Li, Yuan, Ji, and Yan]{zhang2024pcm}
Tao Zhang, Xiangtai Li, Haobo Yuan, Shunping Ji, and Shuicheng Yan.
\newblock Point could mamba: Point cloud learning via state space model.
\newblock \emph{arXiv preprint arXiv:2403.00762}, 2024{\natexlab{a}}.

\bibitem[Zhang et~al.(2024{\natexlab{b}})Zhang, Zhang, and Yan]{zhang2024pcpmae}
Xiangdong Zhang, Shaofeng Zhang, and Junchi Yan.
\newblock Pcp-mae: Learning to predict centers for point masked autoencoders.
\newblock \emph{arXiv preprint arXiv:2408.08753}, 2024{\natexlab{b}}.

\bibitem[Zhang et~al.(2023{\natexlab{b}})Zhang, Gong, Zhang, Li, Qiao, Ouyang, and Yue]{zhang2023metatransformer}
Yiyuan Zhang, Kaixiong Gong, Kaipeng Zhang, Hongsheng Li, Yu Qiao, Wanli Ouyang, and Xiangyu Yue.
\newblock Meta-transformer: A unified framework for multimodal learning.
\newblock \emph{arXiv preprint arXiv:2307.10802}, 2023{\natexlab{b}}.

\bibitem[Zhang et~al.(2021)Zhang, Girdhar, Joulin, and Misra]{zhang2021depthcontrast}
Zaiwei Zhang, Rohit Girdhar, Armand Joulin, and Ishan Misra.
\newblock Self-supervised pretraining of 3d features on any point-cloud.
\newblock In \emph{ICCV}, 2021.

\bibitem[Zhao et~al.(2021)Zhao, Jiang, Jia, Torr, and Koltun]{zhao2021pt}
Hengshuang Zhao, Li Jiang, Jiaya Jia, Philip~HS Torr, and Vladlen Koltun.
\newblock Point transformer.
\newblock In \emph{ICCV}, 2021.

\bibitem[Zheng et~al.(2024)Zheng, Huang, Mei, Hou, Lyu, Dai, Ouyang, and Gong]{zheng2024pointdif}
Xiao Zheng, Xiaoshui Huang, Guofeng Mei, Yuenan Hou, Zhaoyang Lyu, Bo Dai, Wanli Ouyang, and Yongshun Gong.
\newblock Point cloud pre-training with diffusion models.
\newblock In \emph{CVPR}, 2024.

\bibitem[Zhou et~al.(2023)Zhou, Wang, Ma, Liu, Huang, and Wang]{zhou2023uni3d}
Junsheng Zhou, Jinsheng Wang, Baorui Ma, Yu-Shen Liu, Tiejun Huang, and Xinlong Wang.
\newblock Uni3d: Exploring unified 3d representation at scale.
\newblock \emph{arXiv preprint arXiv:2310.06773}, 2023.

\bibitem[Zhu et~al.(2023)Zhu, Yang, Wu, Huang, Zhang, He, He, Zhao, Shen, Qiao, et~al.]{zhu2023ponderv2}
Haoyi Zhu, Honghui Yang, Xiaoyang Wu, Di Huang, Sha Zhang, Xianglong He, Tong He, Hengshuang Zhao, Chunhua Shen, Yu Qiao, et~al.
\newblock Ponderv2: Pave the way for 3d foundataion model with a universal pre-training paradigm.
\newblock \emph{arXiv preprint arXiv:2310.08586}, 2023.

\bibitem[Zhu et~al.(2022)Zhu, Zhu, Li, Wu, Li, Wang, and Dai]{zhu2022uniperceiver}
Xizhou Zhu, Jinguo Zhu, Hao Li, Xiaoshi Wu, Hongsheng Li, Xiaohua Wang, and Jifeng Dai.
\newblock Uni-perceiver: Pre-training unified architecture for generic perception for zero-shot and few-shot tasks.
\newblock In \emph{CVPR}, 2022.

\end{thebibliography}
}

\appendix
\maketitlesupplementary

\section{Additional Experiments}

\subsection{Object Detection Fine-tuning}
For the scene-level object detection task, we leverage UniPre3D to pre-train the backbone of the classical VoteNet~\cite{qi2019votenet} using the ScanNet20~\cite{dai2017scannet} dataset. Subsequently, we fine-tune the model for the ScanNet20 detection task within the MMDetection3D~\cite{mmdet3d2020} framework. As shown in Table~\ref{tab:det}, our UniPre3D significantly outperforms prior methods, achieving state-of-the-art performance and delivering substantial improvements, particularly on the challenging mAP@50 metric. These results strongly validate our claim in the main paper that UniPre3D serves as an efficient and effective unified 3D pre-training approach.

\subsection{Ablations on Reference Views}

We conduct additional ablation studies to evaluate the impact of reference view selection strategies and the number of reference views in scene-level experiments. For these ablations, we pre-train the SparseUNet~\cite{choy2019spunet} model on the ScanNet20~\cite{dai2017scannet} dataset and fine-tune it on the downstream semantic segmentation task on the ScanNet dataset. The mean Intersection over Union (mIoU) results on the validation set are presented in Table~\ref{tab:abl_view}.

\vspace{6pt}
\noindent\textbf{Reference View Restriction.} In the main paper, we discuss imposing a restriction on the perspective gap between reference and rendered images. To explicitly analyze its necessity, we present results in the first two rows of Table~\ref{tab:abl_view}. For the experiments without this restriction, reference and rendering view angles are randomly selected across the scene. The quantitative results demonstrate that applying the restriction enhances both pre-training effectiveness and fine-tuning performance. This improvement can be attributed to the fact that, without the restriction, the supplementary image information becomes too weak and irrelevant, failing to appropriately balance the pre-training task complexity.

\vspace{6pt}

\noindent\textbf{Number of Reference Views.} In our implementation, we select eight reference views to provide supplementary texture and color information from the pre-trained image model. In this ablation study, we examine the effect of varying the number of reference views. Specifically, we conduct experiments with 2, 4, 8, and 12 reference views, with quantitative results indicating that eight is the optimal choice. These findings suggest that the supplementary information should neither be too sparse nor too dense. When the number of reference views is too low, the pre-training task remains overly complex. Conversely, when too many reference views are used, the pre-training task becomes overly simplistic, limiting the ability of the backbone model to learn effectively.

\section{Supplementary Visualizations}
\begin{figure*}[t]
  \centering
  \includegraphics[width=\linewidth]{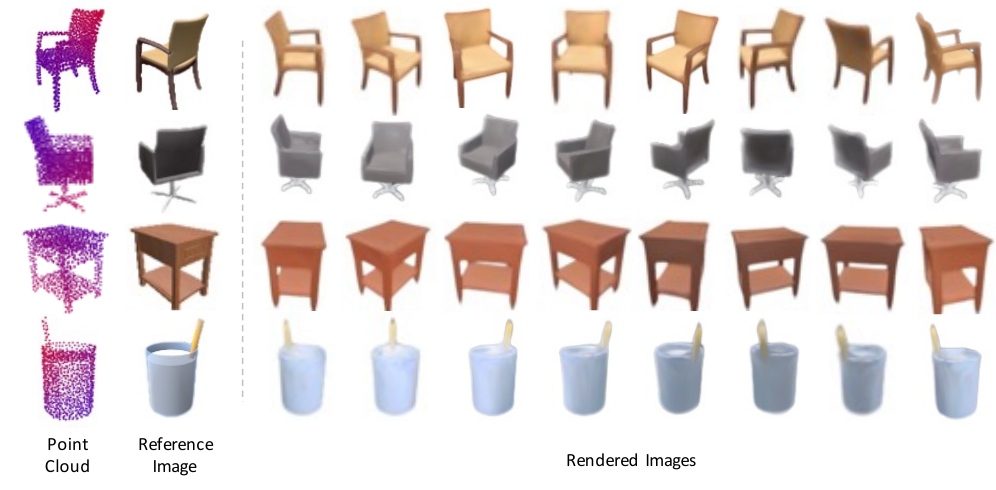}
  \vspace{-20pt}
  \caption{\textbf{Visualization of UniPre3D pre-training outputs on object-level experiments.} The first column presents the input point clouds, followed by the reference view images in the second column. The remaining rows display the rendered images.}
   \label{fig:vis_obj}
\end{figure*}

\begin{figure*}[t]
  \centering
  \includegraphics[width=\linewidth]{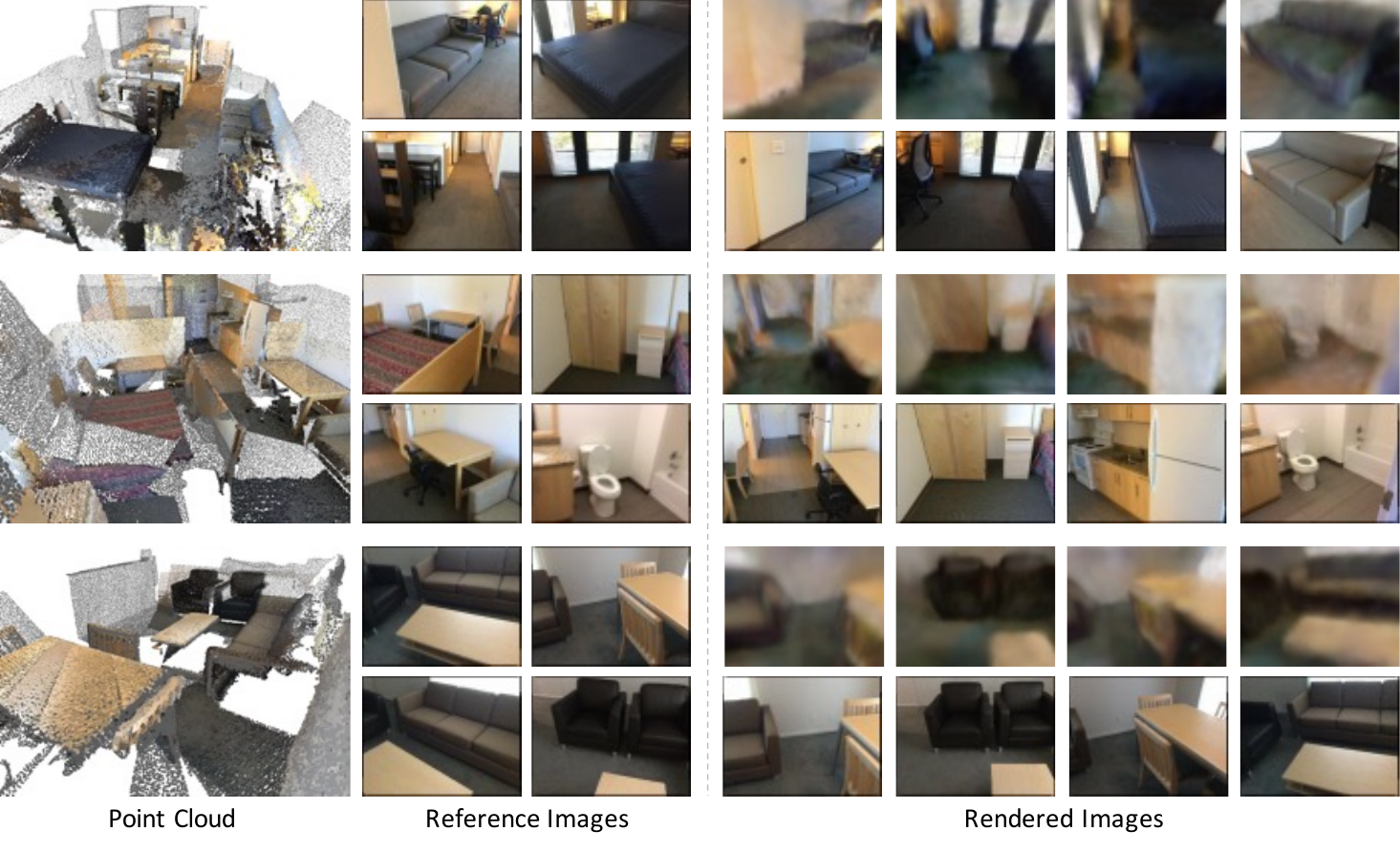}
  \vspace{-20pt}  
  \caption{\textbf{Visualization of UniPre3D pre-training outputs on scene-level experiments.} The first column presents the input point clouds, followed by the reference view images in the second and third columns. The remaining columns display the rendered images (upper rows) and their ground truths (lower rows).}
   \label{fig:vis_scene}
\end{figure*}

Figures~\ref{fig:vis_obj} and \ref{fig:vis_scene} present additional visualization results from the pre-training stage. For each object sample, we provide the original point cloud alongside one reference image, while for each scene sample, we include the original point cloud and two reference images. The rendered outputs comprise multiple images from varying perspectives to comprehensively illustrate the predicted Gaussian primitives. The object samples highlight how color information from a single view is effectively propagated to other views through the learned geometric structures. The scene samples demonstrate that the backbone model successfully captures complex geometric relationships during pre-training, although some details remain blurred due to the limited number of Gaussian primitives.

\begin{table}[!t]
\setlength\tabcolsep{10pt}
\caption{
\textbf{Object detection on the scene-level ScanNet20~\cite{dai2017scannet}.} We report the mean average precision on the validation set.}
\label{tab:det}
\begin{center}
\vspace{-15pt}
\resizebox{1.0\linewidth}{!}{
\begin{tabular}{llcc}
    \toprule[0.95pt]
        Model & Pre-train & mAP@50 & mAP@25\\
    \midrule[0.6pt]
        \multirow{9}{*}{VoteNet~\cite{qi2019votenet}} & \xmark & 33.5 & 58.6 \\
        & RandomRooms~\cite{rao2021randomrooms} & 36.2 & 61.3 \\
        & PointContrast~\cite{xie2020pc} & 38.0 & 59.2 \\
        & PC-FractalDB~\cite{yamada2022pcfractaldb} & 38.3 & 61.9 \\
        & STRL~\cite{huang2021strl} & 38.4 & 59.5 \\
        & DepthContrast~\cite{zhang2021depthcontrast} & 39.1 & 62.1 \\
        & IAE~\cite{yan2023iae} & 39.8 & 61.5 \\
        & Ponder-RGBD~\cite{huang2023ponder} & 41.0 & 63.6 \\
        & \cellcolor{Gray}UniPre3D & \cellcolor{Gray}\textbf{43.3} & \cellcolor{Gray}\textbf{64.0}\\
    \bottomrule[0.95pt]
\end{tabular}}
\end{center}
\vspace{-10pt}
\end{table}

\begin{table}[!t]
\setlength\tabcolsep{10pt}
\caption{
\textbf{Ablation studies on reference view selection and number choices.} We report the PSNR metric for the pre-training stage and mean IoU for the semantic segmentation fine-tuning task on the ScanNet20~\cite{dai2017scannet} dataset. The backbone is SparseUNet~\cite{choy2019spunet}.}
\label{tab:abl_view}
\begin{center}
\vspace{-15pt}
\resizebox{1.0\linewidth}{!}{
\begin{tabular}{cccc}
    \toprule[0.95pt]
        \multicolumn{2}{c}{Reference View} & \multicolumn{2}{c}{Metric Results} \\
    \cmidrule(lr){1-2}\cmidrule(lr){3-4}
         Restrict & Number & Pre-train PSNR & ScanNet20 mIoU \\ 
    \midrule[0.6pt]
        \xmark & 8 & 16.80& 75.04\\
        \cellcolor{Gray}\cmark & \cellcolor{Gray}8 & \cellcolor{Gray}\textbf{16.82} & \cellcolor{Gray}\textbf{75.76} \\
    \midrule[0.6pt]
        \cmark & 2 & 16.81 & 75.37 \\
        \cmark & 4 & 16.80 & 74.95\\
        \cellcolor{Gray}\cmark & \cellcolor{Gray}8 & \cellcolor{Gray}\textbf{16.82} & \cellcolor{Gray}\textbf{75.76} \\
        \cmark & 12 & 16.71 & 75.18\\
    \bottomrule[0.95pt]
\end{tabular}}
\end{center}
\vspace{-10pt}
\end{table}

\end{document}